\def\year{2021}\relax
\documentclass[letterpaper]{article} 
\usepackage{aaai21}  
\usepackage{times}  
\usepackage{helvet} 
\usepackage{courier}  
\usepackage[hyphens]{url}  
\usepackage{graphicx} 
\usepackage{natbib}  
\usepackage{caption} 
\usepackage{booktabs}
\usepackage{nicefrac}
\usepackage{microtype}
\usepackage{amsmath}
\usepackage{amssymb}
\usepackage{amsthm}

\newtheorem*{theorem*}{Theorem}
\newtheorem{proposition}{Proposition}
\newtheorem*{proposition*}{Proposition}

\newtheorem{corollary*}{Corollary}

\newtheorem*{lemma*}{Lemma}

\newcommand{\E}{\mathbb{E}}

\usepackage{bbm}
\usepackage{caption}
\usepackage{subcaption}
\usepackage{footmisc}
\usepackage{mathtools}
\usepackage{multirow}
\usepackage{dsfont}

\usepackage[dvipsnames]{xcolor}
\title{Implicit Kernel Attention}
\author{Kyungwoo Song\textsuperscript{\rm 1}, Yohan Jung\textsuperscript{\rm 2}, Dongjun Kim\textsuperscript{\rm 2}, Il-Chul Moon\thanks{Corresponding author.}\textsuperscript{\rm 2}\\}
\affiliations{\textsuperscript{\rm 1} Department of AI, University of Seoul \\
	\textsuperscript{\rm 2} Department of ISysE, Korea Advanced Institute of Science and Technology (KAIST) \\
	\texttt{kyungwoo.song@uos.ac.kr, \{becre1776,dongjoun57,icmoon\}@kaist.ac.kr} \\
}

\begin{document}
	\maketitle
	\begin{abstract}
		\textit{Attention} computes the dependency between representations, and it encourages the model to focus on the important selective features. Attention-based models, such as Transformer and graph attention network (GAT), are widely utilized for sequential data and graph-structured data. This paper suggests a new interpretation and generalized structure of the attention in Transformer and GAT. For the attention in Transformer and GAT, we derive that the attention is a product of two parts: 1) the RBF kernel to measure the similarity of two instances and 2) the exponential of $L^{2}$ norm to compute the importance of individual instances. From this decomposition, we generalize the attention in three ways. First, we propose implicit kernel attention with an implicit kernel function instead of manual kernel selection. Second, we generalize $L^{2}$ norm as the $L^{p}$ norm. Third, we extend our attention to structured multi-head attention. Our generalized attention shows better performance on classification, translation, and regression tasks.
	\end{abstract}
	\section{Introduction}
	Attention \cite{bahdanau2014neural} is widely utilized to improve model performance as well as to explain the model mechanisms. 
	The attention in Transformer computes the dot-product between query and key, which are the linear projections of hidden features.
	The attention in GAT also utilizes the dot-product between a weight vector and a concatenation of hidden representations.
	It is well known that the dot-product of two vectors is a product of two distinct terms: 1) the cosine of the angle between two vectors, which computes the similarity; and 2) the individual norm of two vectors, which measures the magnitude of each vector.
	This opens a question on how to analyze the attention with the explicitly separated terms of the similarity and the magnitude, and how to generalize the separation in attentions.
	
	We provide a new interpretation of attention, and this interpretation leads to generalized attention by formalizing the attention weight into a multiplication of two terms: similarity and magnitude.
	We derive the explicit separation, so the derivation reveals that the attention in Transformer and GAT is a product of 1) the Radial Basis Function (RBF) kernel between instances and 2) the exponential of $L^{2}$ norm for each instance.
	
	This paper proposes an implicit kernel function that generalizes the RBF kernel function, embedded in the attention.
	Given that attention uses a kernel implicitly, we note that the kernel function measures the similarity under the inductive bias embedded in the kernel. 
	Traditionally, the modelers manually selected kernels with the expert knowledge for a specific domain or a dataset.
	We construct an implicit kernel function instead of an explicit kernel to find a data-adaptive kernel, which provides better task performances under the data context.
	We formulate the implicit kernel function by constructing the spectral density depending on the dataset because a kernel construction can be interpreted as the spectral density estimation \cite{bochner,yaglom}.
	This paper further develops the formulation by introducing multi-head implicit attention with a structured spectral density estimation.
	
	The magnitude term is the second component in the decomposition of attention, and the magnitude separately measures the importance of each instance pair by an exponential of their $L^2$ norms.
	As $L^{2}$ norm in the attention is rigid without considering the dataset property, 
	we extend $L^{2}$ as the $L^{p}$ norm with a hyper-parameter $p$ to control the scale of magnitude terms and attention weight sparsity.
	The $p$ controls the growth rate of the magnitude terms, which are the scale of individual importance. 
	By treating $p$ as a hyper-parameter, we can impose an inductive bias to focus on the relative importance between inputs (similarity) on translation tasks; or the individual importance on each input (magnitude) for the classification task.
	Besides, we find that the decrement of $p$ is eventually related to the sparsity of the attention weights, empirically and theoretically.
	
	In summary, we derive the decomposition of attention as a product of similarity and magnitude.
	Under the decomposition, we generalize the attention with an implicit kernel function and a $L^{p}$ norm. Further, we extend our works to structured multi-head attention with the implicit kernel functions. Our generalized implicit attention is exchangeable with the attention in Transformer and GAT without increasing the algorithmic complexity in order.
	
	\section{Preliminary} 
	\paragraph{Transformer}
	The attention layers in Transformer compute the importance of each feature to consider the semantics of the given sequence.
	Transformer calculates the importance weight by the dot-product of $q_{i}=h_{i}W_Q$ and $k_{j}=h_{j}W_K$ with the hidden feature $h$, the query $q_{i}$, the key $k_{j}$, and linear projection matrix $W_Q$, $W_K$ \cite{vaswani2017attention}. 
	The attention weight $\alpha_{ij}$ in Eq. \ref{eq:scaled_dot_attention} utilizes the softmax function with a scaling $1/\sqrt{d_k}$ where $d_k$ is the dimension of $k_{j}$
	\begin{equation}
	\alpha_{ij}=\frac{\exp\left(q_{i}^{T}k_{j}/\sqrt{d_k}\right)}{\sum_{l}\exp\left(q_{i}^{T}k_{l}/\sqrt{d_k}\right)}
	\label{eq:scaled_dot_attention}
	\end{equation}
	\paragraph{Graph Attention Network}
	GAT \cite{Petar2018gat} adopts the attention in Eq. \ref{eq:graph_attention} to aggregate the relevant neighborhood's features.
	GAT transforms each hidden feature, $h_{i}$ and $h_{j}$, with a linear projection matrix, $W$. GAT aggregates features by concatenating the transformed hidden features. After the concatenation, GAT computes the dot-product between a weight vector, $a$, and aggregated features.
	GAT adopts LeakyReLU \cite{maas2013rectifier}, and it should be noted that LeakyReLU has a different slope parameter, $c$, for the positive region and the negative region. GAT utilizes the softmax function for the neighbor of a node $i$, $N_i$.
	\begin{equation}
	\alpha_{ij}=\frac{\exp\left(\text{LeakyReLU}\left(a^T[Wh_{i}||Wh_{j}]\right)\right)}
	{\sum_{l \in N_{i}}\exp\left(\text{LeakyReLU}\left(a^T[Wh_{i}||Wh_{l}]\right)\right)}
	\label{eq:graph_attention}
	\end{equation}
	\paragraph{Multi-Head Attention}
	Both Transformer and GAT extend the attention to Multi-Head Attention (MHA) to capture the diverse features by introducing the different linear projection matrix. The original MHA assumes the independence between muti-heads, and this paper provides a dependence structure by using the spectral densities of multi-heads.
	
	\subsection{Kernel}
	\paragraph{Spectral Density}
	Due to the dual property given by Bochner theorem \cite{bochner}, a stationary kernel, $k(x,x')$, is selected uniquely by a corresponding spectral density, $p(w)$, and vice versa, as in Eq. \ref{eq:spectral_density}.
	The hypothesis class of the spectral density function determines the adaptiveness of a kernel, and the explicit specification on the density function class could restrict the adaptiveness. Hence, we provide an implicit density function class through an implicit generative model. 
	For instance, the generator in Generative Adversarial Network (GAN) \cite{goodfellow2014generative} is able to represent spectral density, flexibly \cite{ikl2019}.
	Generally, Eq. \ref{eq:spectral_density} with a spectral density, $p(w)$, is intractable, so we approximate Eq. \ref{eq:spectral_density} by the Monte Carlo (MC) integration with $R$ sampled spectral points, $w_{r}$, as shown in Eq. \ref{eq:spectral_aproximation} \cite{ton2018spatial}.
	
	\begin{align}
	k(x,x')&=\int_{\mathcal{R}^{d}}\text{exp}\left(iw^T\left(x-x'\right)\right)p(w)dw \label{eq:spectral_density}\\
	&\approx \frac{1}{R}\sum_{r=1}^{R}\text{exp}\left(iw_{r}^T\left(x-x'\right)\right) \label{eq:spectral_aproximation}
	\end{align}
	
	\paragraph{Kernel in Attention}
	Tsai et al. \cite{tsai2019transformer} utilize a kernel to formulate the attention weights. They replace the $\exp(q_{i}^{T}k_{j}/\sqrt{d_{k}})$ in scaled dot-product attention with a manually selected kernel, such as the RBF kernel, without an explicit decomposition.
	Because they do not derive the decomposition explicitly, they only utilize the RBF kernel for a similarity term without a magnitude term. We name it RBF-only for the comparison.
	
	The concurrent works \cite{katharopoulos2020transformers, choromanski2020rethinking, choromanski2020masked} focus on efficiency and propose linear time complexity attention based on the relationship between kernel and attention. This paper focuses on the new interpretation and generalization of attention to improve the capacity of attention in Transformer and GAT. 
	We emphasize the improved capacity of attention with linear time complexity can be reached by combining our approach with the concurrent works.
	\subsection{Copula}
	The original MHA in Transformer and GAT lacks an explicit model to handle the dependency between multiple attentions.
	Therefore, we formulate a copula-augmented spectral density estimation to consider the dependency between the spectral densities in each head.
	The copula is a cumulative distribution function (CDF) defined on the unit cube with uniform marginals \cite{copula2007nelsen}. Formally, a copula, $C$, is defined as $C\left(u_1,...,u_d\right) = P\left(U_1 \leq u_1,...,U_d \leq u_d\right)$, where the marginal distribution on $U_i$ is uniformly defined on [0,1].
	By Sklar's theorem \cite{sklar1959functions}, we can represent a joint cumulative distribution of $x_1,...,x_d$ as a marginal distribution of each random variables and their copula, $F\left(x_1,...x_d\right)=C\left[F_1(x_1),...F_d(x_d)\right]$.
	We impose the dependencies between attentions in MHA by estimating each head's spectral density jointly with a copula augmented inference.
	\section{Methodology}
	\subsection{Decomposition of Transformer and GAT}
	This section derives the decomposition of attention in Transformer and GAT as the product of two distinct terms: 1) similarity term and 2) magnitude term. 
	The similarity measures the relative importance, and the magnitude computes the individual importance.	
	We derive the decomposition for the scaled dot-product attention in Proposition \ref{eq:remark_transformer}.
	\begin{proposition}
		Let $\alpha_{ij}$ be $\frac{1}{Z_{1}(\alpha)}\exp\left(\frac{q_{i}^{T}k_{j}}{\sqrt{d_k}}\right)$ where $Z_{1}(\alpha)$ is a normalizing constant. Then $\alpha_{ij}$ has the form:
		\begin{equation}
		\centering
		\frac{1}{Z_{1}(\alpha)} \times \underbrace{\exp\left({\frac{-\|q_{i}-k_{j}\|_{2}^{2}}{2\sqrt{d_k}}}\right)}_\text{similarity}\times                        \underbrace{\exp\left(\frac{\|q_{i}\|_{p=2}^{2}+\|k_{j}\|_{p=2}^{2}}{2\sqrt{d_k}}\right)}_\text{magnitude} \nonumber
		\end{equation}
		\label{eq:remark_transformer}
	\end{proposition}
	The scaled dot-product attention has the similarity term, $f_{RBF}=\exp\left({\frac{-\|q_{i}-k_{j}\|_{2}^{2}}{2\sqrt{d_k}}}\right)$, which is the RBF kernel with fixed length-scale hyper-parameters, $\sqrt[\leftroot{-3}\uproot{3}4]{d_k}$.
	Additionally, the scaled dot-product attention yields the magnitude term, $\exp\left(\frac{\|q_{i}\|_{p=2}^{2}+\|k_{j}\|_{p=2}^{2}}{2\sqrt{d_k}}\right)$, and it 
	measures the individual importance of each instance as $\|q_{i}\|_{p=2}^{2}$ and $\|k_{j}\|_{p=2}^{2}$ with $L^{2}$ norm. 
	A large similarity term and a large magnitude term induce the large attention weight for the given pair of query and key.
	
	On GAT, we formulate the decomposition for the attention in Eq. \ref{eq:graph_attention}.
	The decomposition in Proposition \ref{eq:remark_gat} is slightly different from the decomposition of Transformer.
	For an attention weight between node $i$ and node $j$, GAT computes the two similarity terms: 1) the similarity between $q_{i}$ and $ca$, 2) the similarity between $ca$ and $k_{j}$. From the decomposition, we can interpret the weight vector, $a$, as the medium that connects $q_{i}$ and $k_{j}$.
	\begin{proposition}
		Define $\alpha_{ij}= \frac{\exp\left(\text{LeakyReLU}\left(a^T[Wh_{i}||Wh_{j}]\right)\right)}
		{Z_{2}(\alpha)}$, $q_{i}=\left[Wh_{i}||\overrightarrow{0}\right]$, and $k_{j}=\left[\overrightarrow{0}||Wh_{j}\right]$ where $Z_{2}(\alpha)$ is a normalizing constant, and $c$ be the slope parameter in Leaky-ReLU.
		Then $\alpha_{ij}$ has the form:
		\begin{align}
		\centering
		\frac{1}{Z_{2}(\alpha)} 
		&\times
		\underbrace{\exp\left({\frac{-\|q_{i}-ca\|_{2}^{2}}{2}}\right)
			\times \exp\left({\frac{-\|ca-k_{j}\|_{2}^{2}}{2}}\right)}_\text{similarity} \nonumber \\
		& \times \underbrace{\exp\left(\frac{2\|ca\|_{p=2}^{2}+ \|q_{i}\|_{p=2}^{2}+\|k_{j}\|_{p=2}^{2}}{2}\right)}_\text{magnitude} \nonumber
		\end{align}
		\label{eq:remark_gat}
	\end{proposition}
	
	\begin{figure*}[t]
		\centering
		\begin{subfigure}{.13\textwidth}
			\centering
			\includegraphics[width=\linewidth]{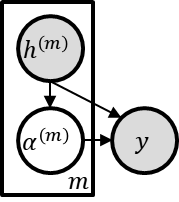}
			\caption{Attention}
			\label{fig:att}
		\end{subfigure}\hfil
		\begin{subfigure}{.20\textwidth}
			\centering
			\includegraphics[width=\linewidth]{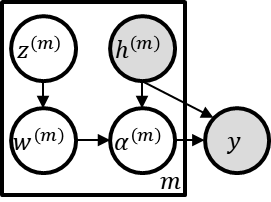}
			\caption{$\text{IKA}_{(\text{S},\text{NS})}$, IKAN}
			\label{fig:ika} 
		\end{subfigure}\hfil
		\begin{subfigure}{.19\textwidth}
			\centering
			\includegraphics[width=\linewidth]{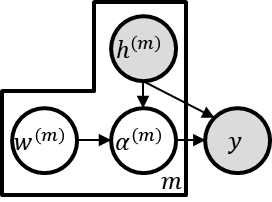}
			\caption{IKAN-direct}
			\label{fig:ikandirect}
		\end{subfigure}\hfil
		\begin{subfigure}{.19\textwidth}
			\centering
			\includegraphics[width=\linewidth]{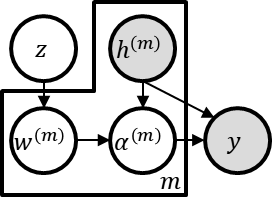}
			\caption{MIKAN}
			\label{fig:mikan}
		\end{subfigure}
		\begin{subfigure}{.20\textwidth}
			\centering
			\includegraphics[width=\linewidth]{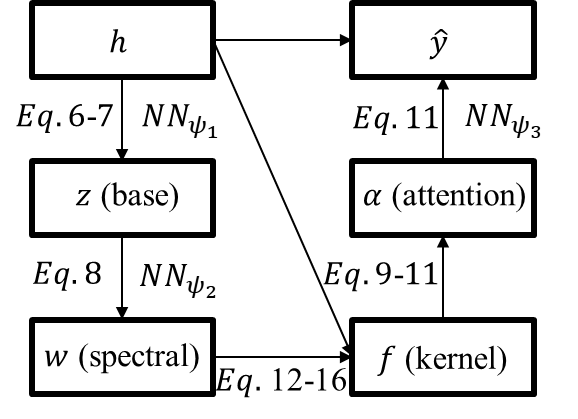}
			\caption{Structure overview}
			\label{fig:network}
		\end{subfigure}
		\caption{Visualization of the attention and our models.
			To capture the spectral density flexibly, we first sample $z$ from the base distribution $p(z)$ and transform it to $w$ with a flexible function such as a neural network.
			The $\text{IKA}_{\text{S}}$, $\text{IKA}_{\text{NS}}$ and IKAN in (b) estimate the base distribution and the spectral density in each head depending on the dataset, unlike the scaled dot-product attention. 
			IKAN-direct in (c) optimizes spectral points $w$ directly, and MIKAN in (d) estimates the base distribution $p(z^{(1:M)})$ jointly.  We provide a structure overview of our model in (e).}
		\label{fig:pgm(att)}
	\end{figure*}
	\subsection{Variations of Implicit Kernel Attention}
	From the factorization, we propose a new attention method, which formulates an implicit kernel function, and which utilizes a $L^{p}$ norm.
	Attention in Transformer and GAT uses a fixed single Gaussian spectral density, $p(w)$, and its corresponding RBF kernel in Figure \ref{fig:att}.
	In contrast, we propose implicit kernel attention (IKA) that estimates the spectral density implicitly to find an appropriate kernel depending on the dataset in Figure \ref{fig:ika}.
	Second, we interpret $p$ in $L^{p}$ norm as a hyper-parameter, so we define IKA with norm (IKAN).
	Besides, we propose a deterministic and simple but effective model, IKAN-direct, to optimize spectral points $w$, directly in Figure \ref{fig:ikandirect}.
	Lastly, we propose multi-head IKAN (MIKAN), which adopts a copula-augmented inference to estimate a structured spectral density of MHA, jointly in Figure \ref{fig:mikan}.
	Figure \ref{fig:network} represents the structure overview of our models.	
	
	\subsection{IKA: Implicit Kernel Attention}
	RBF measures the similarity by the Euclidean distance.
	Recent works \cite{li2019geodesic,chen2018metrics} conjecture that the Euclidean metric might not be proper in the hidden feature space.
	Furthermore, RBF includes the exponential term, and the range of exponential is a positive real-value without zero.
	Therefore, the sparsity cannot be achieved in the attention weight because of the positive nature.
	
	By following \cite{bochner, yaglom}, the implicit spectral density allows us to construct an implicit kernel, so this paper proposes learning an implicit spectral density from the dataset.
	To formulate an implicit spectral density, we construct a base distribution, $p(z)$, such as a Gaussian distribution, and we sample $z$ from $p(z)$. Then, we transform the sampled $z$ with a flexible function such as a neural network to construct the spectral density, $p(w)$. It should be noted that the spectral density should be symmetric for the real-valued kernel, which we add constraints in Eq. \ref{eq:w_m}.
	
	Under the implicit spectral density, the log marginal likelihood, $\log p(y|h)$, is intractable where $y$ is an output variable and $h$ is a hidden feature. Therefore, we derive the evidence lower bound (ELBO) of the log marginal likelihood with an inference network, $q(z|h)$. We maximize the ELBO in Eq. \ref{eq:IKAN_ELBO} with a re-parameterization method to backpropagate the gradients \cite{KingmaW13}.
	\begin{align}
	\mathcal{L}=\E_{q(z|h)}\left[\log p(y,z|h)\right] - \E_{q(z|h)}\left[\log q(z|h)\right]
	\label{eq:IKAN_ELBO}
	\end{align}
	
	For MHA in Transformer or GAT, we independently sample $z$ from $q(z|h)$ for each head by formulating $q(z|h) = \prod_{m=1}^{M}q(z^{(m)}|h^{(m)})$.
	We construct each inference network, $q(z^{(m)}|h^{(m)})$, with a neural network parameterized by $\psi_{1}$ as in Eq. \ref{eq:q}.
	We formulate $z^{(m)}$ as a concatenation of $\widetilde{z}^{(m)}$ and $-\widetilde{z}^{(m)}$ to preserve the symmetric of a base distribution.
	\begin{align}
	\mu^{(m)},\log\sigma^{(m)} &= NN_{\psi_{1}}(h^{(m)}) \label{eq:q} \\
	\widetilde{z}^{(m)} \sim \mathcal{N}(\mu^{(m)},(\sigma^{(m)})^{2}\mathrm{I}),\quad &
	z^{(m)}=\left[\widetilde{z}^{(m)} || - \widetilde{z}^{(m)} \right]
	\label{eq:z}
	\end{align}
	
	A symmetric base distribution enables us to construct an unconstrained neural network $NN_{\psi_{2}}$ to transform $z$ into $w$.
	After sampling $z^{(m)}$, we transform it with $NN_{\psi_{2}}$ to construct the spectral density in Eq. \ref{eq:w_m}.
	To preserve the symmetricity of the spectral density, we use the absolute value of $|z^{(m)}|$ as an input, and we multiply the sign of $z^{(m)}$ \cite{ikl2019}.
	\begin{align}
	&w^{(m)}=\text{sign}\left(z^{(m)}\right) \times NN_{\psi_{2}}\left(|z^{(m)}|\right) \label{eq:w_m} 
	\end{align}
	Eq. \ref{eq:ika_transformer} and Eq. \ref{eq:ika_gat} denote the formulation of un-normalized attention weights, $\widetilde{\alpha}_{ij}^{(m)}$, with the implicit kernel function $f$ for Transformer and GAT, respectively.
	From the un-normalized attention weights, $\widetilde{\alpha}_{ij}^{(m)}$; we compute attention weights, $\alpha_{ij}^{(m)}$, with normalization in Eq. \ref{eq:y_pred}.
	The implicit kernel function, $f$, is a similarity term of attention in Proposition \ref{eq:remark_transformer} and \ref{eq:remark_gat}. 
	The rest of the structure is the same as the original Transformer and GAT with a neural network $NN_{\psi_3}$ in Eq. \ref{eq:y_pred}.
	\begin{align}
	&\widetilde{\alpha}_{ij}^{(m)} = f\left(q_i^{(m)},k_j^{(m)},w^{(m)}\right) \times \exp(\frac{\|q_{i}^{(m)}\|_{2}^{2}+\|k_{j}^{(m)}\|_{2}^{2}}{2\sqrt{d_k}}) \label{eq:ika_transformer} \\
	&\widetilde{\alpha}_{ij}^{(m)} = f\left(q_i^{(m)},ca^{(m)},w^{(m)}\right) \times f\left(ca^{(m)},k_j^{(m)},w^{(m)}\right) \nonumber \\
	&\quad \quad \times \exp(\frac{2\|ca^{(m)}\|_{2}^{2}+\|q_{i}^{(m)}\|_{2}^{2}+\|k_{j}^{(m)}\|_{2}^{2}}{2}) \label{eq:ika_gat} \\
	&\widehat{y} = NN_{\psi_{3}}\left(\alpha^{(1:M)},h^{(1:M)}\right),
	\quad \alpha_{ij}^{(m)} = \frac{\widetilde{\alpha}_{ij}^{(m)}}{\sum_{l}\widetilde{\alpha}_{il}^{(m)}} \label{eq:y_pred}
	\end{align}
	\paragraph{$\text{IKA}_{\text{S}}$: Implicit Kernel Attention with Stationary Kernel}
	This section explains the construction of kernel function $f$, and we omit the multi-head related notation, $m$, for simplicity.
	We approximate a continuous stationary kernel with the MC integration with $R$ sampled spectral points, $w_{r}$, and its random Fourier feature map, $\phi_{r}$, by taking the real part in Eq. \ref{eq:spectral_aproximation} \cite{ton2018spatial,rahimi2008random,jung2020approximate}.
	\begin{gather}
	f(q_i,k_j,w) = \frac{1}{R}\sum_{r=1}^{R}\phi_r(q_i)^T\phi_r(k_j) \label{eq:bochner1}\\
	\phi_r(q_i) = \binom{\text{cos}\left(w_r^Tq_i\right)}{\text{sin}\left(w_r^Tq_i\right)}, \phi_r(k_j) = \binom{\text{cos}\left(w_r^Tk_j\right)}{\text{sin}\left(w_r^Tk_j\right)} \label{eq:bochner2}
	\end{gather}
	
	We generalize the attention by replacing the RBF in the scaled dot-product attention with an implicit kernel function, $f$ as shown in Eq. \ref{eq:ika_transformer}. We name this attention as $\text{IKA}_{\text{S}}$. 
	Similarly, we can construct the implicit kernel, $f$, for GAT in Eq. \ref{eq:ika_gat} by following Eq. \ref{eq:bochner1} and \ref{eq:bochner2}.
	
	In spite of the above kernel construction, the output of the kernel function in Eq. $\ref{eq:bochner1}$ might have a negative value, so it contradicts with the non-negative property of attention weights. As a result, we square Eq. \ref{eq:bochner1}, $f^2$, to ensure the non-negativity of IKA.
	Proposition \ref{propr:square} claims that the squared implicit kernel function still generalizes the RBF kernel that is utilized in the scaled dot-product attention.
	\begin{proposition}
		Let $k_{RBF}^{(l)}$ be the RBF kernel function with a lengthscale $l$, and $f$ be its approximation with $R$ random Fourier features by Eq. \ref{eq:bochner1} and \ref{eq:bochner2}.  If we sample spectral points, $w$, from a Gaussian distribution $ \mathcal{N}\left(0,\frac{1}{2l^{2}}\mathrm{I}\right)$, then $lim_{R \rightarrow \infty} f^{2} = k_{RBF}^{(l)}$.
		\label{propr:square}
	\end{proposition}
	
	\paragraph{$\text{IKA}_{\text{NS}}$: Implicit Kernel Attention with Non-Stationary Kernel}
	Similar to the stationary kernel, we can approximate a continuous non-stationary kernel with the random Fourier feature map $\phi_{r}$ in Eq. \ref{eq:yaglom1}-\ref{eq:yaglom2} \cite{ton2018spatial,yaglom}.
	We formulate a new attention, $\text{IKA}_{\text{NS}}$, induced by a non-stationry kernel function in Eq. \ref{eq:yaglom1}-\ref{eq:yaglom2}.
	Different from stationary kernel, we sample two types of $R$ spectral points, $w_{1,r}$ and $w_{2,r}$, from two distinct implicit spectral density estimators.
	Similarly, we construct an implicit kernel, $f$, for GAT in Eq. \ref{eq:ika_gat} by following Eq. \ref{eq:yaglom1}-\ref{eq:yaglom2}.
	\begin{gather}
	f(q_i,k_j,w) = \frac{1}{4R}\sum_{r=1}^{R}\phi_r(q_i)^T\phi_r(k_j) \label{eq:yaglom1}\\
	\phi_r(q_i) = \binom{\text{cos}(w_{1,r}^Tq_i)+\text{cos}(w_{2,r}^Tq_i)}{\text{sin}(w_{1,r}^Tq_i)+\text{sin}(w_{2,r}^Tq_i)}
	\\
	\phi_r(k_j) = \binom{\text{cos}(w_{1,r}^Tk_j)+\text{cos}(w_{2,r}^Tk_j)}{\text{sin}(w_{1,r}^Tk_j)+\text{sin}(w_{2,r}^Tk_j)} \label{eq:yaglom2}
	\end{gather}
	
	The error bound for a non-stationary kernel approximation is not well explored. To validate $\text{IKA}_{\text{NS}}$, we show that the sufficient enough samples from a spectral density guarantee the non-stationary kernel approximation in Proposition \ref{remark:errorbound}.
	\begin{proposition}
		Let $k(x,x')$ be a non-stationary kernel, $f(x,x')$ be an approximated kernel with $R$ sampled spectral points, $w_{1}$ and $w_{2}$, 
		and $\mathcal{M}$ be a compact subset of $\mathcal{R}^{d}$ with a diameter D. Let $\sigma_{1}^{2}=\E[w_{1}^Tw_{1}]$, $\sigma_{2}^{2}=\E[w_{2}^Tw_{2}]$ and $\mathcal{E}=sup_{x,x' \in \mathcal{M}} |f(x,x')-k(x,x')|$. Then, \\
		i) $P\left[\mathcal{E} \geq \epsilon \right] 
		\leq 
		2^{8} \times \left(\frac{D\sqrt{\sigma_{1}^{2}+\sigma_{2}^{2}}}{\epsilon}\right)^{2} \times \exp\left(\frac{-R\epsilon^{2}}{2(d+1)}\right)$ \\
		ii) $\mathcal{E} \leq \epsilon$ with any constant probability when $R=\Omega(\frac{d}{\epsilon^{2}} \log{\frac{D\sqrt{\sigma_{1}^{2}+\sigma_{2}^{2}}}{\epsilon}} )$ 
		\label{remark:errorbound}
	\end{proposition}
	Similar to the $\text{IKA}_{\text{S}}$, we use $f^{2}$ instead of $f$ to ensure the non-negativity of the attention weights.
	
	\subsection{IKAN: Implicit Kernel Attention with $L^{p}$}
	\paragraph{IKAN}
	In addition to the implicit kernel in $\text{IKA}_{\text{NS}}$, we generalize the $L^{2}$ norm in the magnitude term in Proposition \ref{eq:remark_transformer} and \ref{eq:remark_gat} to be the $L^p$, which determines the importance of individual representations.
	The optimal $p$ might be different for each dataset and task, so we select $p$ through experiments.
	There are tasks where the similarity is more important than the magnitude and vice versa. Treating $p$ as a hyper-parameter that determines the scale of magnitude term imposes an inductive bias on the model.
	Furthermore, we analyze the relationship between the magnitude and the sparsity of attention weights. 
	For the analysis, we define attention weights, $\alpha_{ij}(p)$, whose magnitude utilizes $\|x\|_{p}=\left(|x_{1}|^{p}+...+|x_{d}|^{p}\right)^{1/p}$ instead of $\| x\|_{2}$. It should be noted that $\|x\|_{p}$ is an absolutely homogeneous function for $p > 0$.
	We show that $\alpha_{ij}(p)$ becomes sparse when $p$ goes to zero in Proposition \ref{eq:pnorm}.
	
	\begin{proposition}
		Let $\mathds{1}_{A_{p}^{(i)}}$ be the multi-dimensional indicator function with $n$-th component to be 1 if $n\in A_{p}^{(i)}$ and 0 otherwise, where $A_{p}^{(i)}$ is the set of index for maximal elements: $A_{p}^{(i)}=\{l| l=argmax_{j}\alpha_{ij}(p)\}$. Then, $\lim_{p \to 0^{+}} \alpha_{ij}(p) = \mathds{1}_{A_{p}^{(i)}}/|A_{p}^{(i)}|$
		\label{eq:pnorm}
	\end{proposition}
	\paragraph{IKAN-direct} 
	We propose a simple and effective alternative that sets spectral points, $w$, as a learnable parameter, such as a weight in a neural network.
	IKAN-direct optimizes spectral points, $w$, directly without sampling. It corresponds to handling the spectral density as a mixture of delta distribution.
	\subsection{MIKAN: Multi-Head Attention with IKAN}
	The dependency modeling between heads in MHA is necessary to diversify the attention weights.
	Similar to the individual calculation of the attention in Transformer and GAT, the variations of IKA estimate the spectral density in each head independently.
	Therefore, we introduce $q(z|h)=c_{q}\left(F_{1}(z^{(1)}),,,,,F_{M}(z^{(M)})\right)\prod_{m=1}^{M}q(z^{(m)}|h^{(m)})$, where $c_{q}$ is a copula density; $F_{m}$ is a CDF of each $z^{(m)}$; M is the number of attention head. 
	This alternative variational distribution of $z$, $q(z|h)$, develops IKA to be structured as Multi-head IKAN (MIKAN) by introducing the joint structure through $c_q$.  
	MIKAN maximizes ELBO in Eq. \ref{eq:IKAN_ELBO} with a Monte Carlo estimation by sampling from $q(z|h)$ \cite{KingmaW13}. 
	MIKAN estimates the base distribution without the mean-field assumption by introducing a copula-augmented posterior. IKAN is a special case of MIKAN if we fix $c_{q}$ as a uniform distribution \cite{tran2015copula}. $c_{q}$ can be any copula density, and we set $c_{q}$ as a Gaussian copula with a covariance, $\Sigma$.
	\subsection{Model Complexity}
	Table \ref{table:complexity} denotes the time complexity of computing the attention weights, $\alpha_{ij}$, from a hidden representation, $h$, where $T$ is the length of a sequence or the number of nodes.
	\begin{table}[h!]
		\centering
		\small
		\begin{tabular}{c l l}
			\toprule
			{} & {\bf Model} & {\bf Complexity} \\ \midrule
			\multirow{2}{*}{Baselines} & Transformer & $O(T^{2}d+Td^{2})$ \\
			& RBF-only & $O(T^{2}d+Td^{2})$ \\
			\midrule
			\multirow{7}{*}{Ours} & Expsin & $O(T^{2}d+Td^{2})$ \\
			& Linear & $O(T^{2}d+Td^{2})$ \\
			& $\text{IKA}_{\text{S}}$ & $O(T^{2}R+Td^{2}+TdR)$ \\
			& $\text{IKA}_{\text{NS}}$ & $O(T^{2}R+Td^{2}+TdR)$ \\
			& IKAN & $O(T^{2}R+Td^{2}+TdR)$ \\
			& IKAN-direct & $O(T^{2}R+Td^{2}+TdR)$ \\
			& MIKAN & $O(T^{2}R+Td^{2}+TdR+dM^{3})$ \\
			\bottomrule
		\end{tabular}
		\captionof{table}{
			Time Complexity for computing attention weights where $T, d, R, M$ denotes the length of sequence, hidden dimension, the number of sample size, and the number of attention heads respectively.}
		\label{table:complexity}
	\end{table}
	
	If we consider a task with a long sequence with a high $T$, there is no asymptotic complexity increment in order, so all variations of IKA and baselines fall into the complexity of $O(T^2)$. Also, it should be noted that the complexity of $O(T^{2}R)$ is controllable because $R$ is the number of samples determined by a modeler.
	\section{Results}
	\begin{table*}[ht!]
		\begin{center}
			\begin{tabular}{l|c|c c c c c}
				\toprule
				{\bf Model} & {\bf $\#$ Param.} & {\bf MR} & {\bf CR} & {\bf SST} & {\bf SUBJ} & {\bf TREC} \\ \midrule
				Transformer$^{*}$ & - & 73.1\% & 76.2\% & 76.1\% & 86.3\% & 83.4\% \\
				\midrule
				Transformer & 27.9M & 74.45$\pm$0.94\% & 77.98$\pm$3.16\% & 76.25$\pm$1.60\% & 90.32$\pm$0.68\% & 84.12$\pm$0.77\% \\
				RBF-only & 27.9M & 73.98$\pm$0.48\% & 78.25$\pm$3.02\% & 76.00$\pm$1.82\% & 89.82$\pm$1.16\% & 84.24$\pm$0.90\% \\
				\midrule
				Expsin & 27.9M  & 73.81$\pm$0.98\% & 78.78$\pm$3.26\% & 76.99$\pm$1.21\%  &  89.10$\pm$0.28\% & 84.76$\pm$0.86\%  \\
				Linear & 27.9M & 73.70$\pm$0.49\% & 78.57$\pm$2.86\%  & 76.67$\pm$1.12\%  & 89.22$\pm$0.85\%  & 84.76$\pm$1.41\%  \\
				\midrule
				$\text{IKA}_{\text{S}}$ & 28.1M & 76.06$\pm$0.48\% & 78.83$\pm$3.26\% & 77.37$\pm$1.68\% & 90.66$\pm$0.87\% & 84.88$\pm$1.29\% \\
				$\text{IKA}_{\text{NS}}$ & 28.2M & 76.42$\pm$0.82\% & 79.05$\pm$2.26\% & 77.49$\pm$1.84\% & 90.86$\pm$0.60\% & 85.32$\pm$0.94\%  \\
				IKAN & 28.2M  &  76.63$\pm$0.89\% & 79.21$\pm$3.05\% & 77.49$\pm$1.84\% & \textbf{91.48}$\pm$0.61\% & 85.84$\pm$0.43\% \\
				IKAN-direct & 28.3M & 75.97$\pm$0.54\% & 79.26$\pm$3.05\% & 77.10$\pm$1.68\% & 90.98$\pm$0.53\% & 85.16$\pm$0.38\% \\
				MIKAN & 28.2M & \textbf{76.70}$\pm$1.03\% & \textbf{80.63}$\pm$3.24\% & \textbf{77.69}$\pm$1.85\% & 90.86$\pm$0.61\% & \textbf{85.92}$\pm$0.44\% \\
				\bottomrule
			\end{tabular}
		\end{center}
		\caption{Accuracy for the text classification task. $^{*}$ denotes the reported performance in \cite{complexword20}. $\#$ Param. denotes the number of parameters for the MR dataset. 
		}
		\label{table:text}
	\end{table*}
	\subsection{Sentence Classification}
	We compare our models with the scaled dot-product attention in Transformer \cite{vaswani2017attention}, and RBF-only \cite{tsai2019transformer} on five popular datasets \cite{kim2014convolutional,complexword20}. 
	We perform ten-fold cross validations by following the experimental settings \cite{complexword20}.
	RBF-only represents the scaled dot-product attention with a RBF kernel without a magnitude term.
	To validate the performance for the different kernels, we provide additional models, Expsin and Linear.
	Expsin and Linear denote the scaled dot-product attention that uses periodic kernel and linear kernel, instead of RBF kernel, respectively. For the magnitude term, Expsin and Linear utilize $L^{2}$ norm that is the same as the scaled dot-product attention.
	
	Table \ref{table:text} indicates that the appropriate kernel is different for each dataset.
	The RBF kernel in Transformer performs better than other baselines on MR and SUBJ. 
	On the other hand, Expsin shows better performance on CR and SST, and Linear is superior to RBF on TREC.
	However, $\text{IKA}_{\text{S}}$ and $\text{IKA}_{\text{NS}}$ show consistently better performance than RBF-only, Expsin, and Linear. 
	This consistent tendency demonstrates the importance of data-adaptive implicit kernel empirically, instead of a manual kernel selection.
	Compared to $\text{IKA}_{\text{NS}}$, IKAN shows better performances, and this supports the importance of adapting the magnitude with $p$ as a hyper-parameter.
	IKAN-direct, which optimizes $w$ directly without sampling, shows lower performance than IKAN, but it still performs better than the scaled dot-product attention in Transformer.
	MIKAN with copula augmentation performs best except the SUBJ dataset. 
	
	The second column of Table \ref{table:text} represents the number of parameters for each model on the MR dataset.
	Expsin and Linear have the same parameters as the scaled dot-product in Transformer. For the implicit kernel-based model, we adopt an implicit spectral density estimator, and it requires additional parameters. Because we utilize the shared spectral density estimator for each layer, the implicit kernel-based models require a marginal increment in the number of parameters.
	
\begin{figure}[h]
		\centering
		\begin{subfigure}{0.23\textwidth}
			\includegraphics[width=\textwidth]{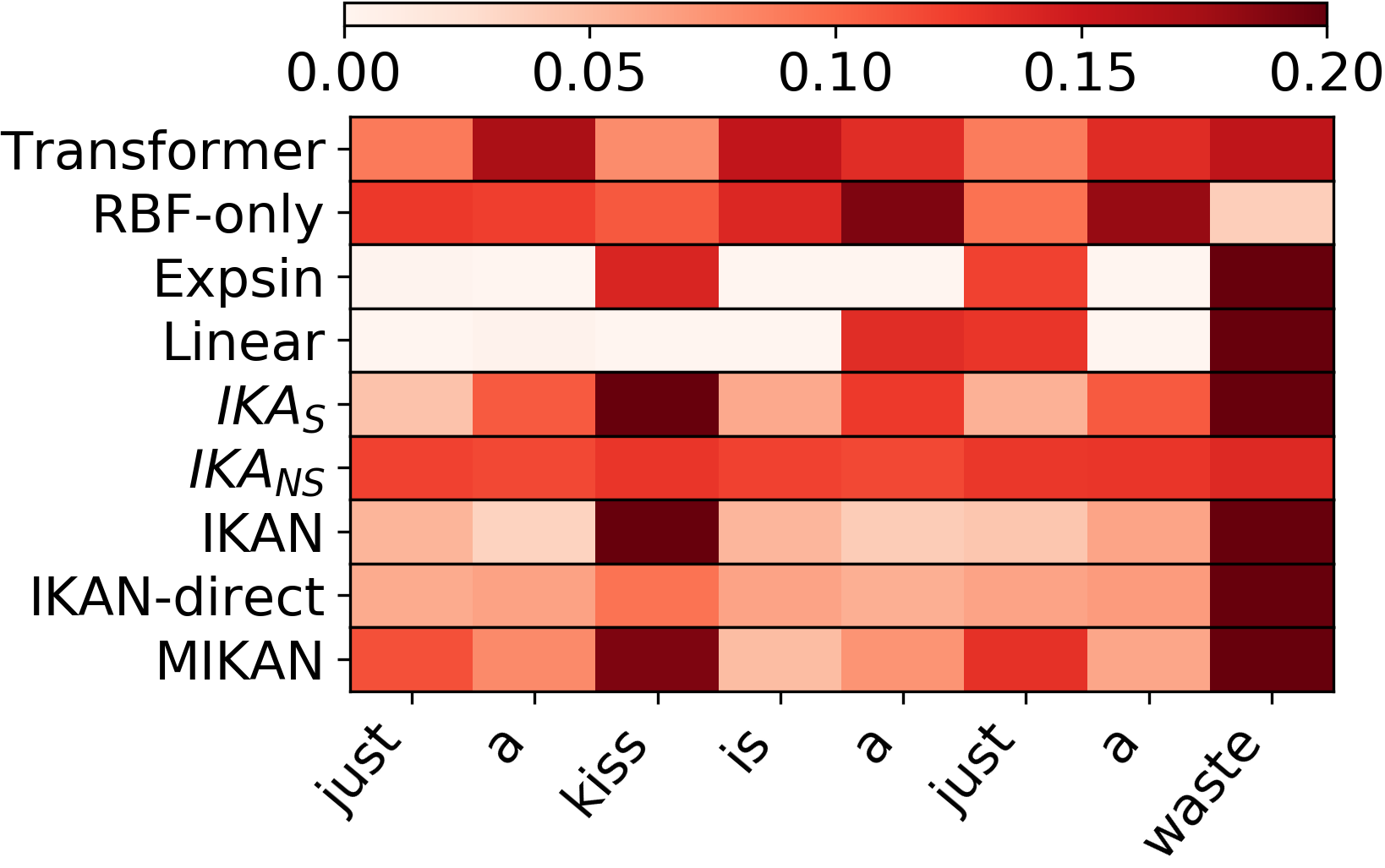}
			\caption{}
			\label{fig:text(1)}
		\end{subfigure}
		\begin{subfigure}{0.23\textwidth}
			\includegraphics[width=\textwidth]{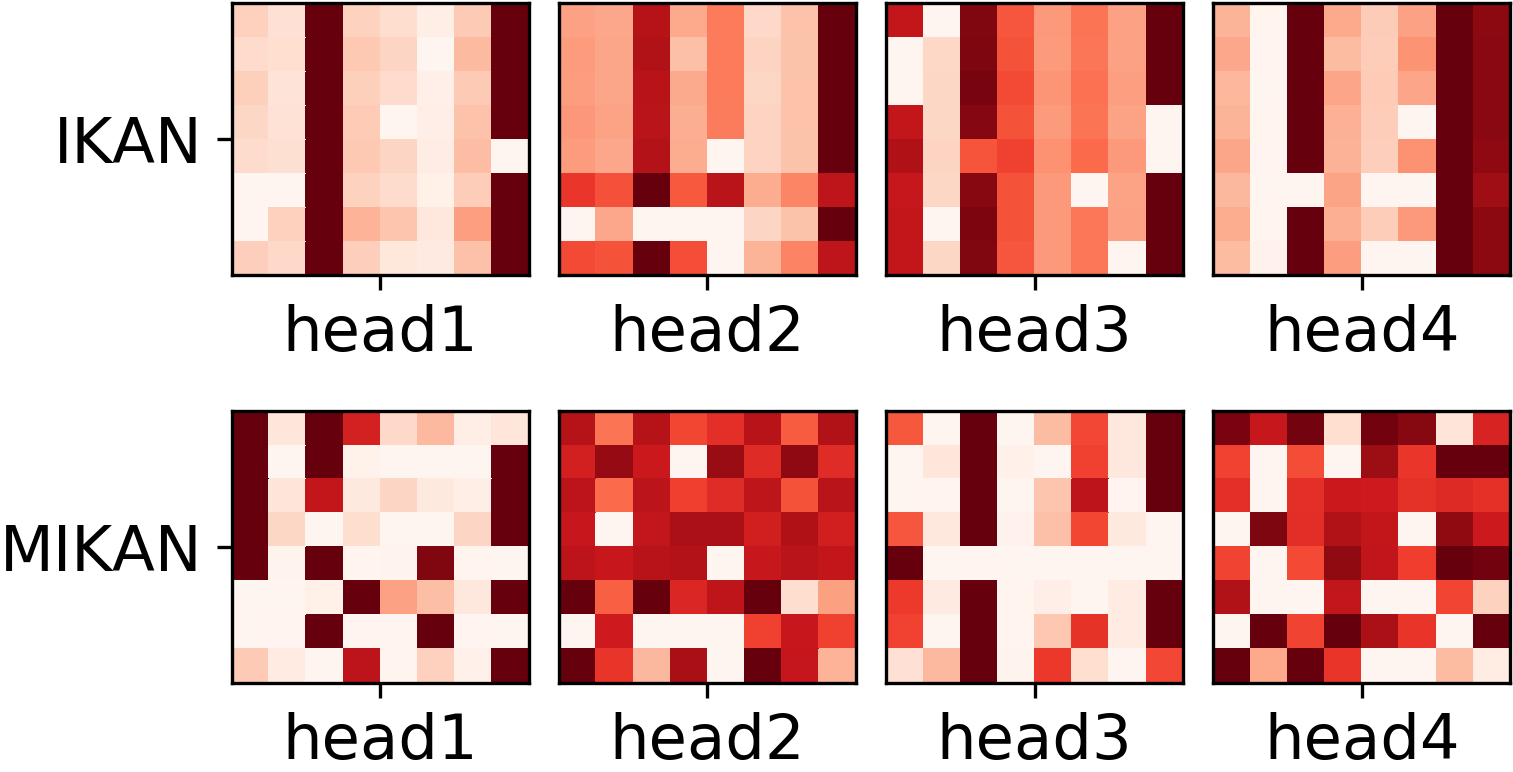}
			\caption{}
			\label{fig:text(2)}
		\end{subfigure}
		\caption{
			(a): Attention weights for the given sentence on MR, whose task is the sentiment classification of a movie review. 
			The sentiment of the given sentence "just a kiss is a just a waste" is negative.
			To identify the sentiment, the model should focus on the \textit{just}, \textit{kiss}, \textit{just}, and \textit{waste}, and MIKAN captures all important words.
			(b): Attention weights matrix of first four heads among eight heads for IKAN and MIKAN.
			MIKAN shows relatively diverse attention weights across the multi-head because of the structured model with the copula.
		}
		\label{fig:text}
	\end{figure}

	Figure \ref{fig:text(1)} and \ref{fig:text(2)} show attention weights for a sentence in MR. In Figure \ref{fig:text(1)}, our implicit kernel-based attention captures the important words \textit{kiss} and \textit{waste}. Besides, MIKAN focuses on the additional important word, \textit{just}.
	Figure \ref{fig:text(2)} represents the attention weights of the first four heads among eight heads in Transformer.
	Each row and column of the square matrix corresponds to the individual word in the given sentence.
	All first four heads in IKAN focus on the same last word \textit{waste}. The first, second, and fourth heads in IKAN also focus on the words \textit{kiss} together.
	On the other hand, MIKAN that adopts structured MHA shows the relatively diverse attention weights across the multi-heads.
	
	\subsection{Translation}
	We compare our models and baselines on IWSLT14 De-En.
	We include MIKAN and IKAN-direct because MIKAN is a generalized model of IKA and IKAN.
	Table \ref{table:translation} shows the results of three runs for Transformer, IKAN-direct and MIKAN. Our models perform better than other models, Beam Search Optimization \cite{wiseman2016sequence}, Actor-Critic \cite{BahdanauBXGLPCB17}, Neural PBMT + LM \cite{jang2017categorical}, Minimum Risk Training \cite{edunov2018classical}, Variational Attention \cite{deng2018latent}, and Transformer.

\begin{figure}[h!]
		\centering
		\begin{subfigure}{.20\textwidth}
			\centering
			\includegraphics[width=\linewidth]{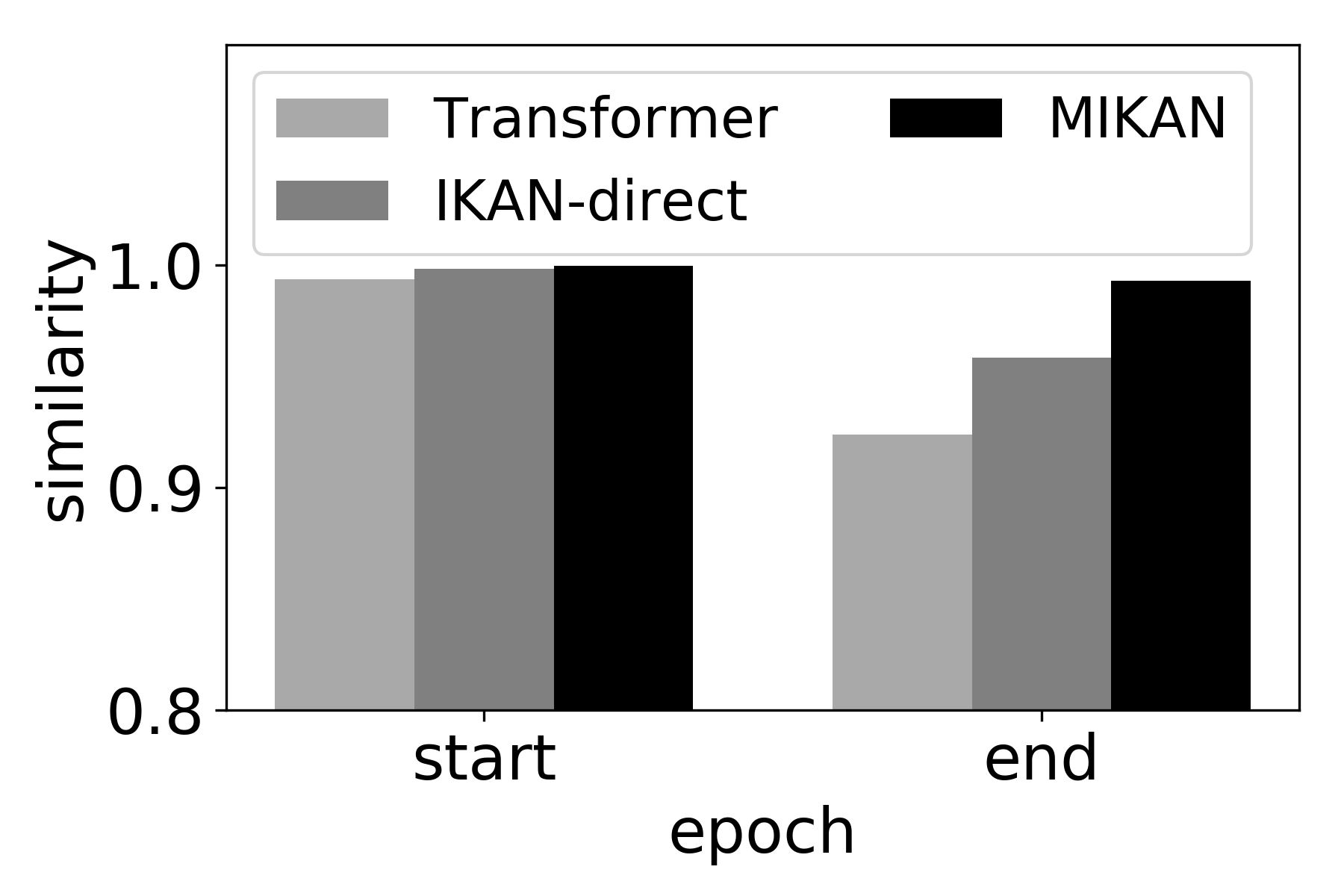}
			\caption{Similarity in self.}
			\label{fig:encoder(scores)}
		\end{subfigure}
		\begin{subfigure}{.20\textwidth}
			\centering
			\includegraphics[width=\linewidth]{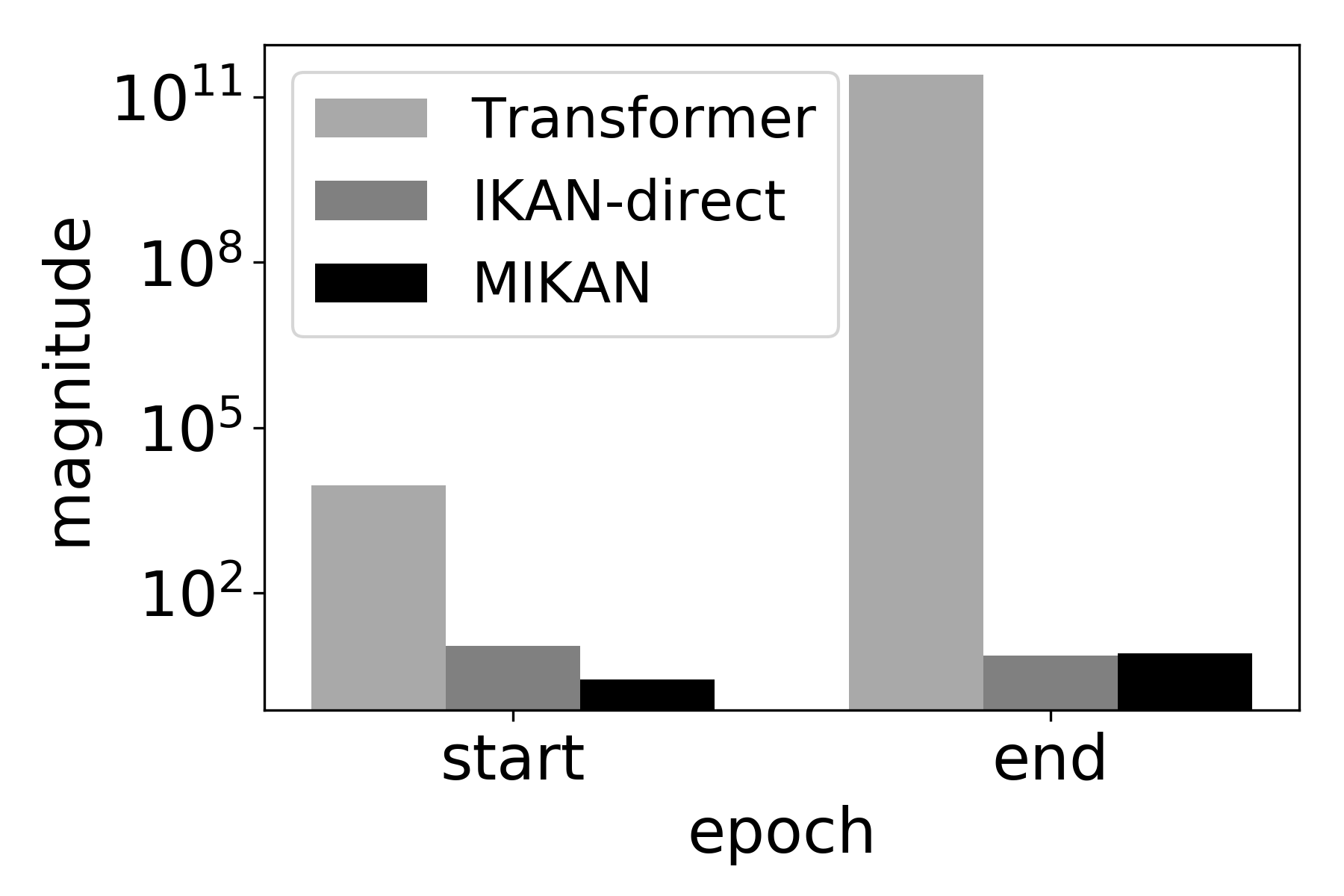}
			\caption{Magnitude in self.}
			\label{fig:encoder(lpnorm)}
		\end{subfigure}
		\hfill
		\begin{subfigure}{.20\textwidth}
			\centering
			\includegraphics[width=\linewidth]{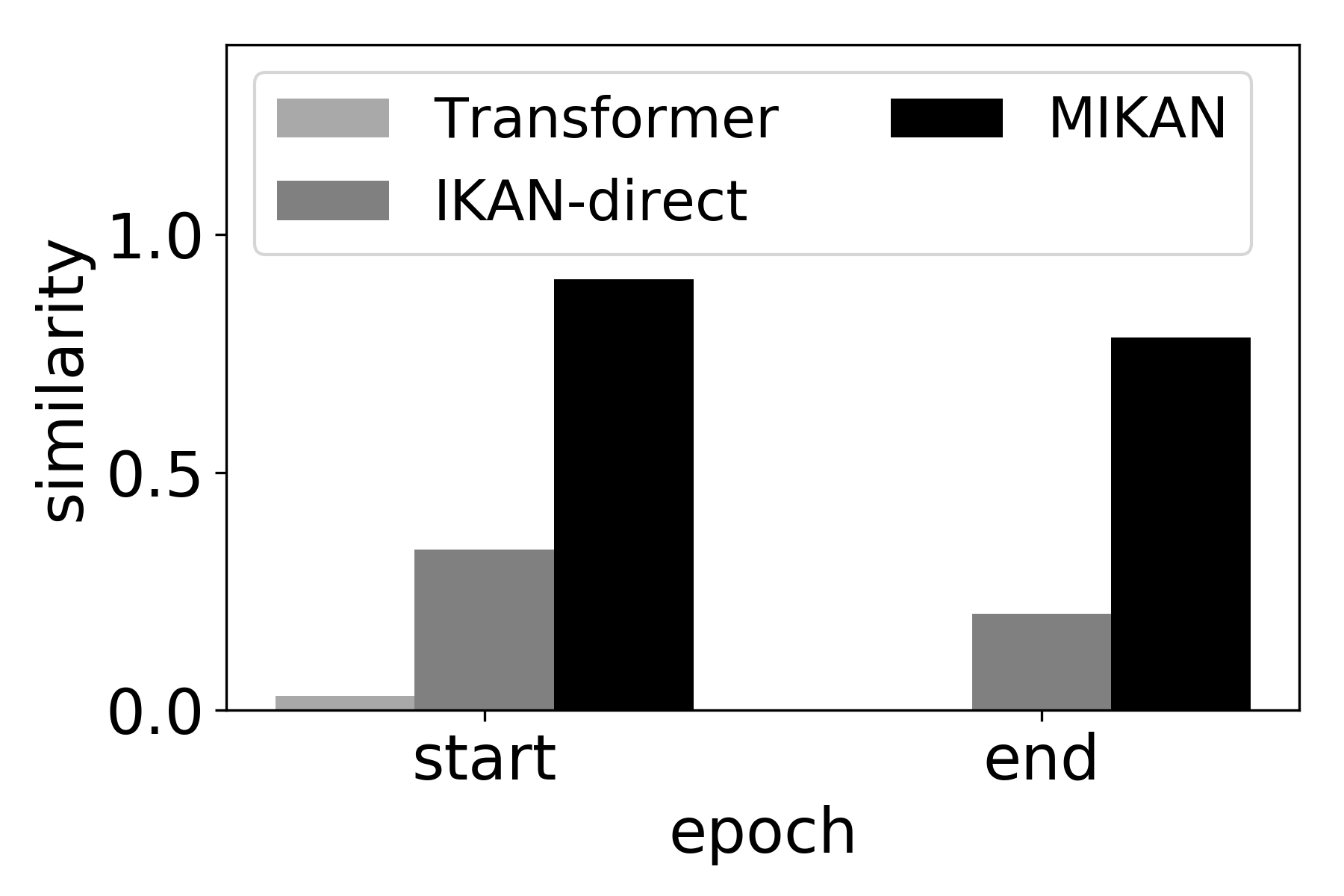}
			\caption{Similarity in cross.}
			\label{fig:cross(scores)}
		\end{subfigure}
		\begin{subfigure}{.20\textwidth}
			\centering        \includegraphics[width=\linewidth]{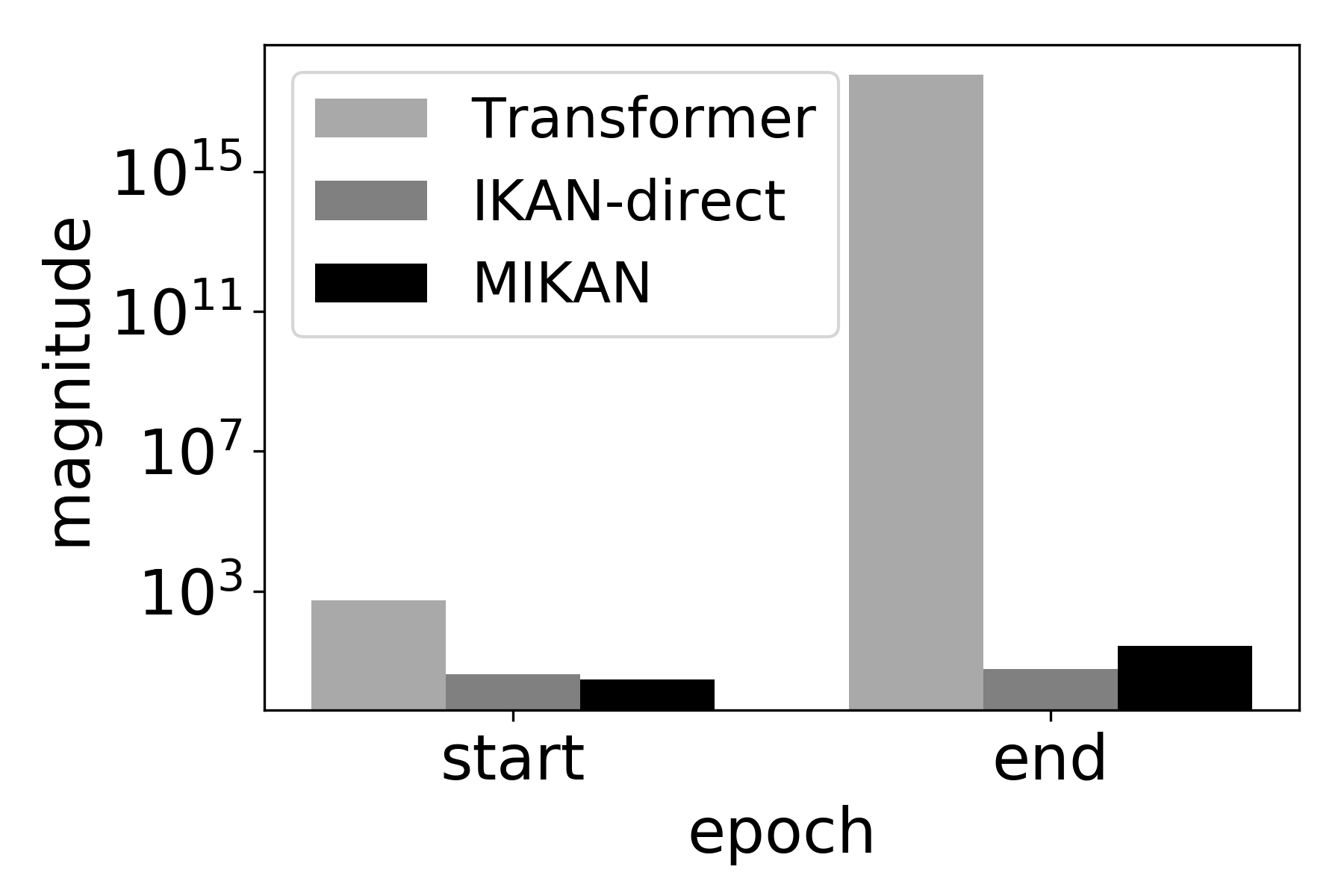}
			\caption{Magnitude in cross.}
			\label{fig:cross(lpnorm)}
		\end{subfigure}
		\caption{Similarity and magnitude of encoder self-attention layer (self.) and decoder-encoder attention layer (cross.).}
		\label{fig:translation}
	\end{figure}

	We analyze the changes in similarity and magnitude terms in Figure \ref{fig:translation}. 
	Figure \ref{fig:encoder(scores)} and \ref{fig:encoder(lpnorm)} show the change in the encoder self-attention layer (self.); and Figure \ref{fig:cross(scores)} and \ref{fig:cross(lpnorm)} represent the changes in the decoder-encoder cross attention layer (cross.). 
	We visualize the changes of maximum similarity and average magnitude from the first epoch (start) to the last epoch (end).
	Figure \ref{fig:encoder(lpnorm)} and \ref{fig:cross(lpnorm)} show that Transformer heavily depends on the magnitude term, instead of similarity. 
	We see the same phenomena in the cross attention layer, where we hypothesize that the similarity is important for the alignment given the pair-wise dependency.
	This phenomenon occurs at the start of training, and it gets worse during the training.
	On the other hand, IKAN and MIKAN represent the stable scale of similarity and magnitude because of the implicit kernel and the $L^{p}$ norm. 
	\begin{table}[h]
		\centering
		\begin{tabular}{l c}
			\toprule
			{\bf Model} & {\bf BLEU} \\ \midrule
			Beam Search Optimization$^{*}$  & 26.36 \\
			Actor-Critic$^{*}$  & 28.53 \\
			Neural PBMT + LM$^{*}$  & 30.08 \\
			Minimum Risk Training$^{*}$ & 32.84 \\
			Variational Attention$^{*}$ & 33.69 \\
			Transformer & 34.44 $\pm$ 0.07 \\
			\midrule
			IKAN-direct & 34.59 $\pm$ 0.09 \\
			MIKAN & \textbf{34.70} $\pm$ 0.09 \\
			\bottomrule
		\end{tabular}
		\captionof{table}{BLEU for IWSLT14 De-En. $^{*}$ denotes the reported performance in \cite{deng2018latent}.}
		\label{table:translation}
	\end{table}
	
	\begin{table*}[t!]
		\begin{center}
			\begin{tabular}{l c c c c c}
				\toprule
				{\bf Model} & {\bf CO2} & {\bf Passenger} & {\bf Housing} & {\bf Concrete} & {\bf Parkinsons} \\ \midrule
				GP (RBF) & 0.060 $\pm$ 0.003 & 0.174 $\pm$ 0.052 & 0.234 $\pm$ 0.045 & 0.178 $\pm$ 0.012  & 0.075 $\pm$ 0.006 \\
				GP (SM)& 0.057 $\pm$ 0.002 & 0.171 $\pm$ 0.048 & 0.191 $\pm$ 0.044 & 0.170 $\pm$ 0.021 & 0.072 $\pm$ 0.004 \\
				Transformer & 0.057 $\pm$ 0.002 & 0.153 $\pm$ 0.043 & 0.126 $\pm$ 0.038 & 0.120 $\pm$ 0.017 & 0.040 $\pm$ 0.005 \\ \midrule
				IKAN-direct & {\bf 0.055} $\pm$ 0.002 & {\bf 0.110} $\pm$ 0.039 & 0.118 $\pm$ 0.038 & 0.111 $\pm$ 0.018 &  {\bf 0.018} $\pm$ 0.004 \\
				MIKAN & {\bf 0.055} $\pm$ 0.002 &  0.142 $\pm$ 0.044 & {\bf 0.108} $\pm$ 0.029 & {\bf 0.108} $\pm$ 0.015 & 0.037 $\pm$ 0.004  \\
				\bottomrule
			\end{tabular}
		\end{center}
		\caption{RMSE on the UCI Regression dataset.}
		\label{table:regression}
	\end{table*}
	\subsection{Regression}
	We apply Transformer on the UCI regression dataset, and we replace the scaled dot-product attention with our models. We perform 10-fold cross-validations, and Table \ref{table:regression} reports the performance following \cite{tompkins2019black}. According to the result, our models perform better than other baselines. 
	
\begin{figure}[h]
		\centering
		\begin{subfigure}[b]{.32\columnwidth}
			\includegraphics[width=\textwidth]{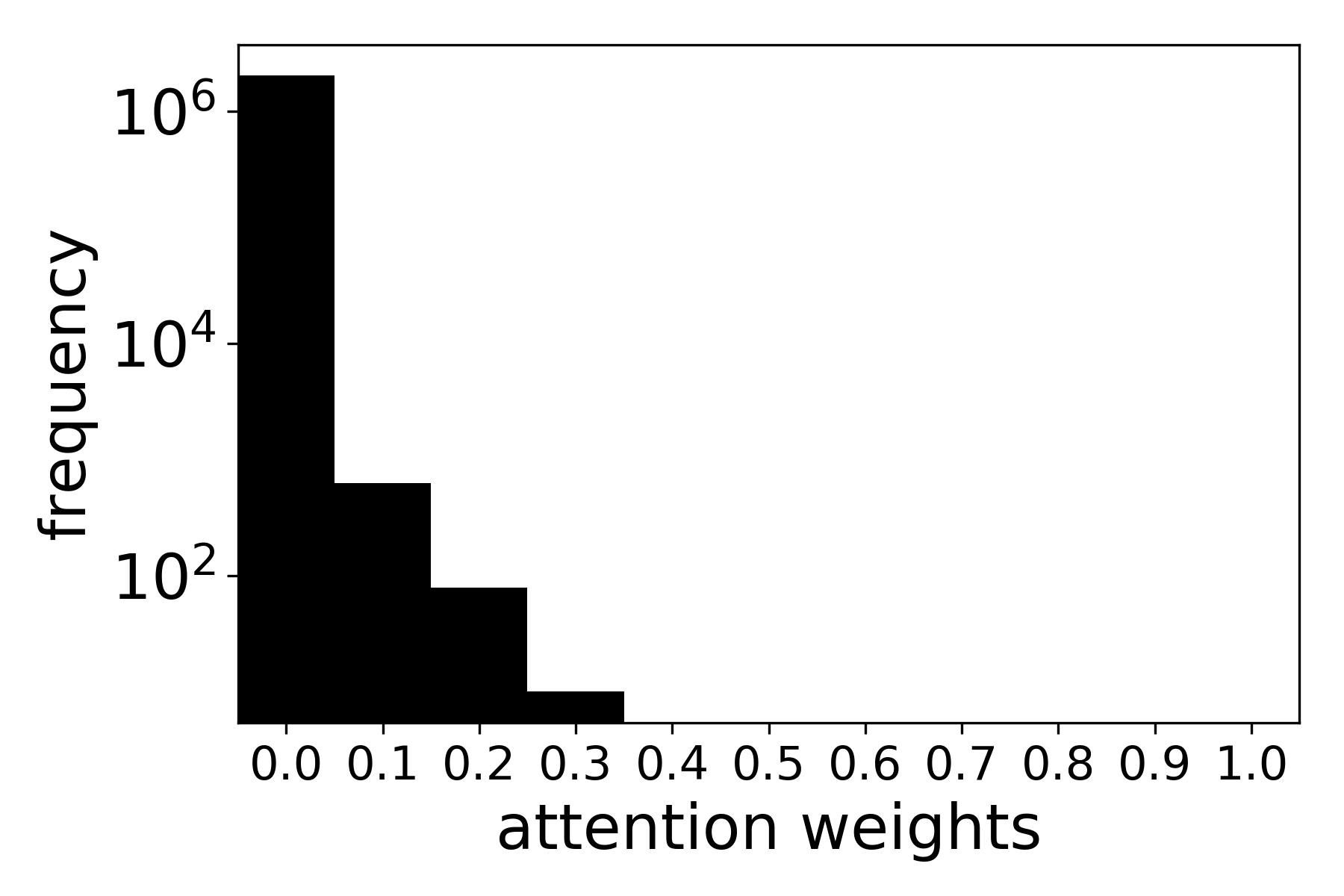}
			\caption{$p=2.0$}
			\label{fig:mikan_pnorm1}
		\end{subfigure}
		\begin{subfigure}[b]{.32\columnwidth}
			\includegraphics[width=\textwidth]{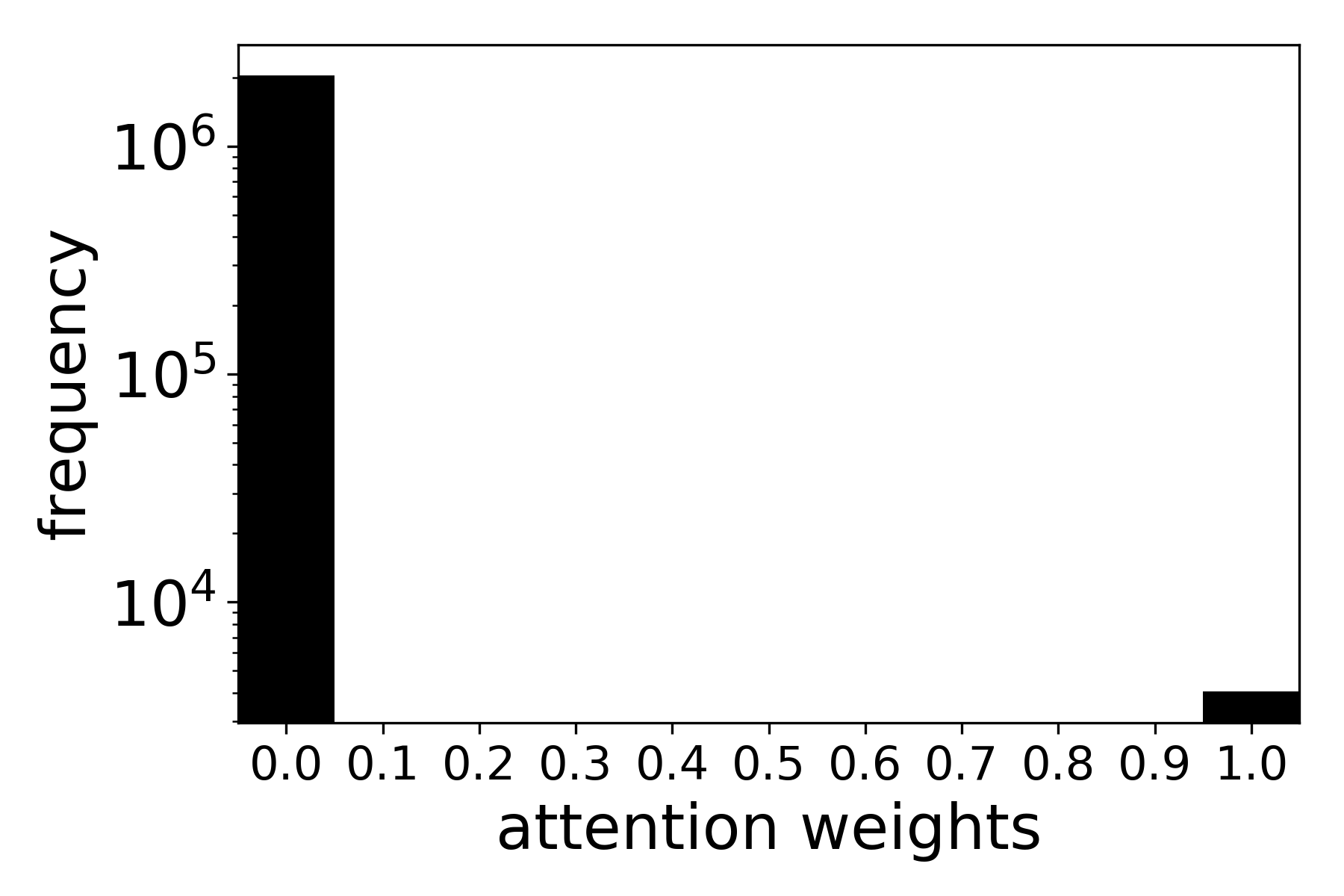}
			\caption{$p=0.1$}
			\label{fig:mikan_pnorm3}
		\end{subfigure}
		\begin{subfigure}[b]{.32\columnwidth}
			\includegraphics[width=\textwidth]{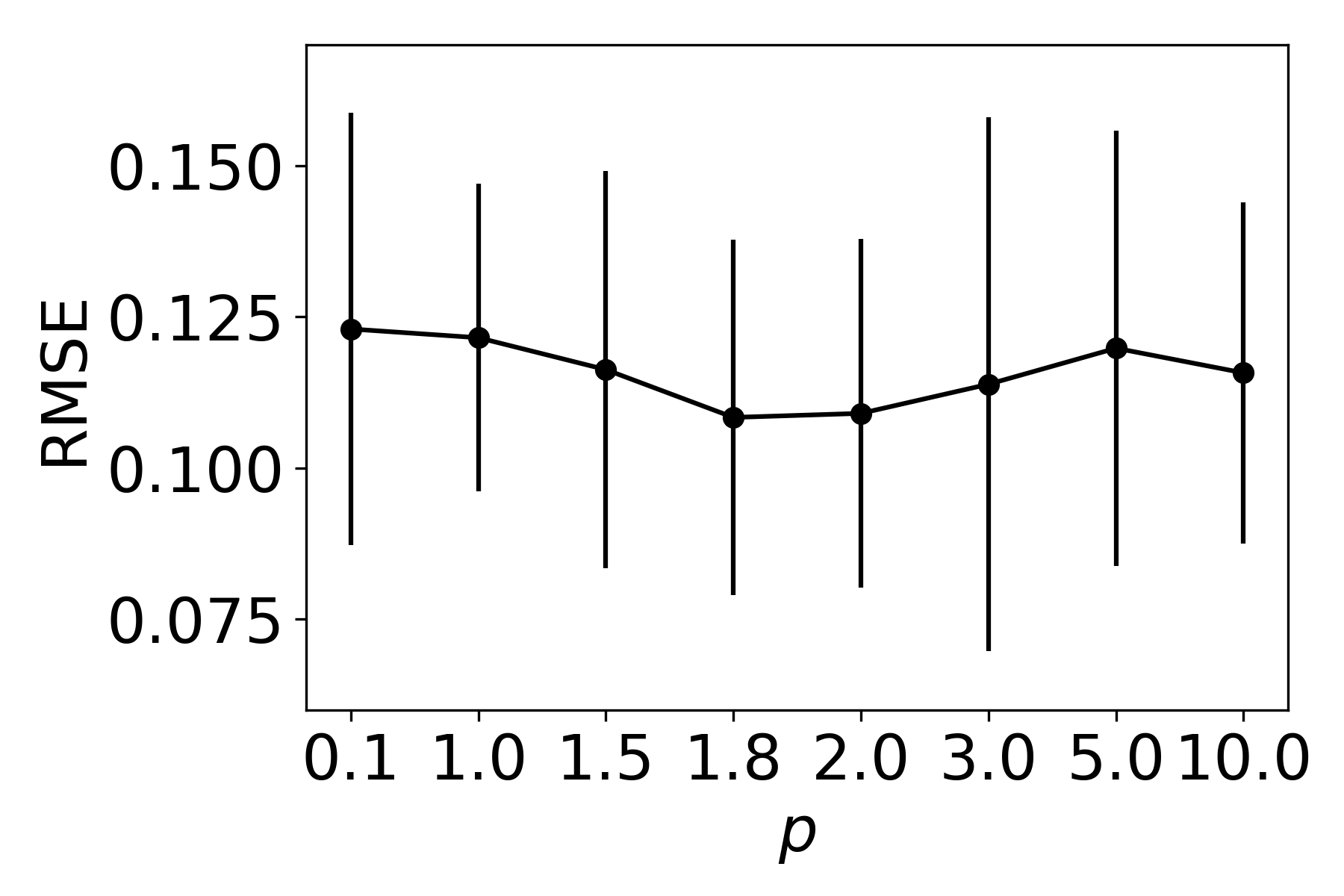}
			\caption{Sensitivity}
			\label{fig:mikan_pnorm_sensitivity}
		\end{subfigure}
		\caption{(a), (b): A histogram for attention weights in MIKAN for different $p$. (c) shows that RMSE of MIKAN varies depending on $p$.}
		\label{fig:regression}
	\end{figure}

	Afterward, we perform the qualitative analysis to clarify the relationship between the $p$ and the sparsity of attention weights.
	Figure \ref{fig:mikan_pnorm1} and \ref{fig:mikan_pnorm3} visualize a histogram of MIKAN attention weights for different $p$s on the Housing dataset. As $p$ goes to zero, most attention weights have either zero or one. Figure \ref{fig:mikan_pnorm_sensitivity} shows the sensitive analysis with respect to $p$. The RMSE varies depending on $p$, but all results show better performance than the baseline.
	\subsection{Node Classification}
	We compare GAT and our generalized attention in Proposition \ref{eq:remark_gat}.
	We perform ten-fold cross validations, and Table \ref{table:graph} represents the average accuracy for DeepWalk \cite{perozzi2014deepwalk}, ICA \cite{lu2003link}, Chebyshev \cite{defferrard2016convolutional}, GCN \cite{gcnKipfW17}, GAT \cite{Petar2018gat}.
	\begin{table}[h!]
		\small
		\centering
		\begin{tabular}{lccc}
			\toprule
			{\bf Model} & {\bf Cora} & {\bf Citeseer} & {\bf Pubmed}\\ 
			\midrule
			DeepWalk$^{*}$ & 67.2\% & 43.2\% & 65.3\%\\
			ICA$^{*}$ & 75.1\% & 69.1\% & 73.9\%\\
			Chebyshev$^{*}$ & 81.2\% & 69.8\% & 74.4\%\\
			GCN$^{*}$ & 81.5\% & 70.3\% & 79.0\% \\
			GAT$^{*}$ & 83.0$\pm$0.7\% & 72.5$\pm$0.7\% & 79.0$\pm$ 0.3\%\\
			\midrule
			IKAN-direct & 83.3$\pm$0.7\% & 72.8$\pm$1.0\% & \textbf{79.2}$\pm$0.4\%\\
			MIKAN & \textbf{83.4}$\pm$0.6\% &\textbf{72.9}$\pm$0.8\%&  79.1$\pm$0.6\% \\
			\bottomrule
		\end{tabular}
		\captionof{table}{Accuracy for the node classification task. $^{*}$ denotes the reported performance in \cite{Petar2018gat}.}
		\label{table:graph}
	\end{table}
	\begin{figure}
		\centering
		\begin{subfigure}{.40\columnwidth}
			\includegraphics[width=\textwidth]{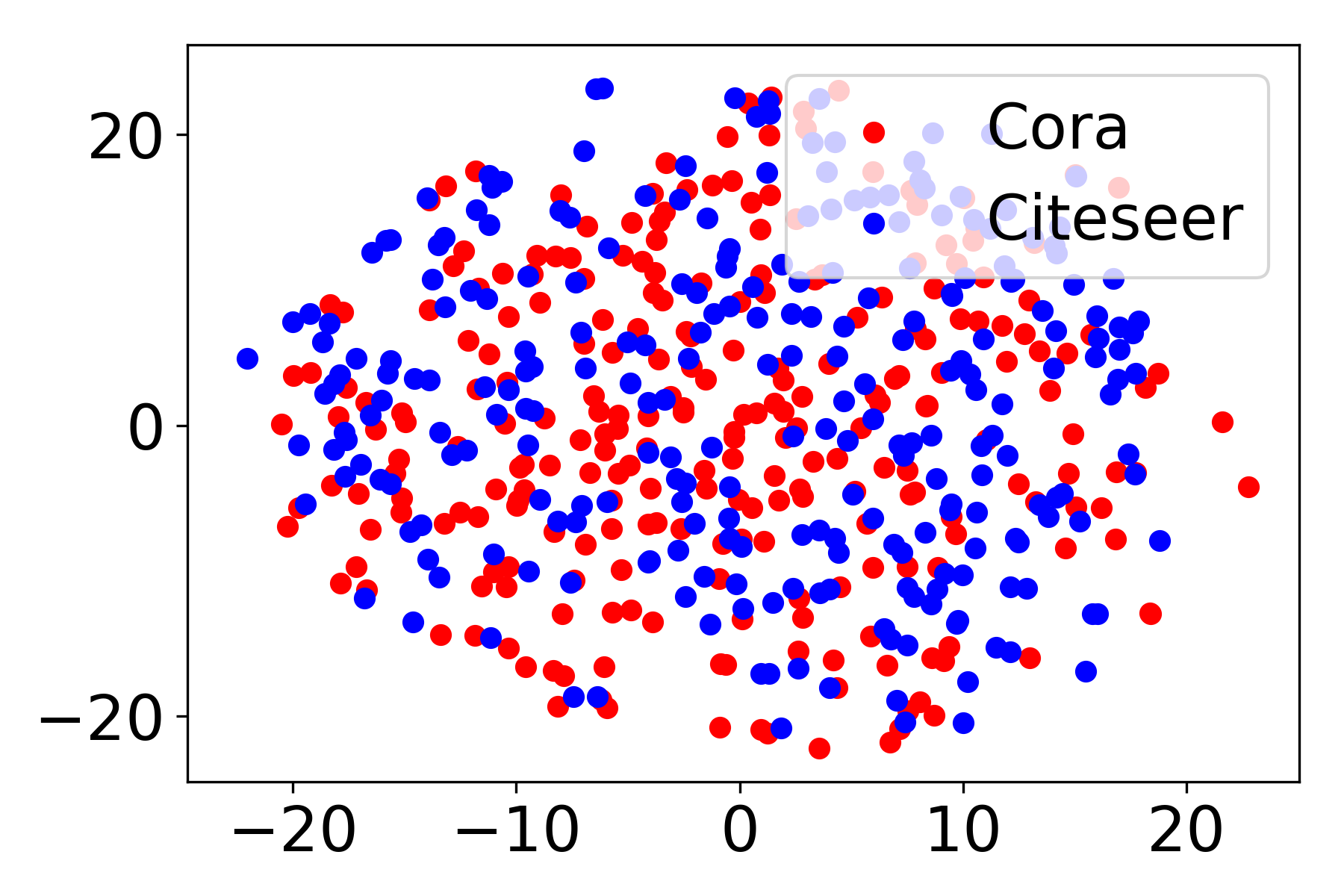}
			\caption{$w$ in GAT}
			\label{fig:graph_rbfw}
		\end{subfigure}
		\mbox{}\hspace{1.4em}
		\begin{subfigure}{.40\columnwidth}
			\includegraphics[width=\textwidth]{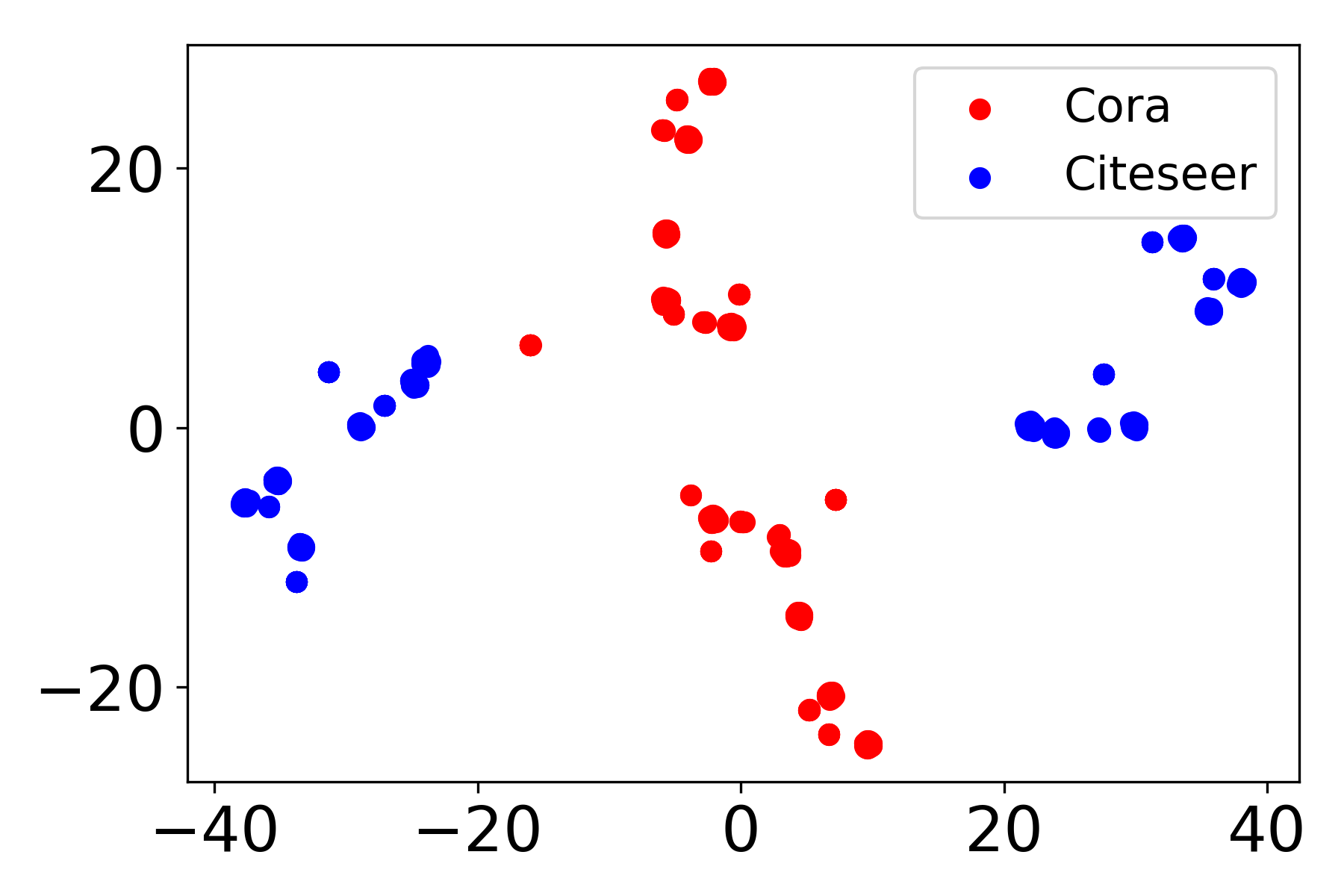}
			\caption{$w_{1}$ in MIKAN}
			\label{fig:graph_mikanw1}
		\end{subfigure}
		\hfill
		\begin{subfigure}{.45\columnwidth}
			\includegraphics[width=\textwidth]{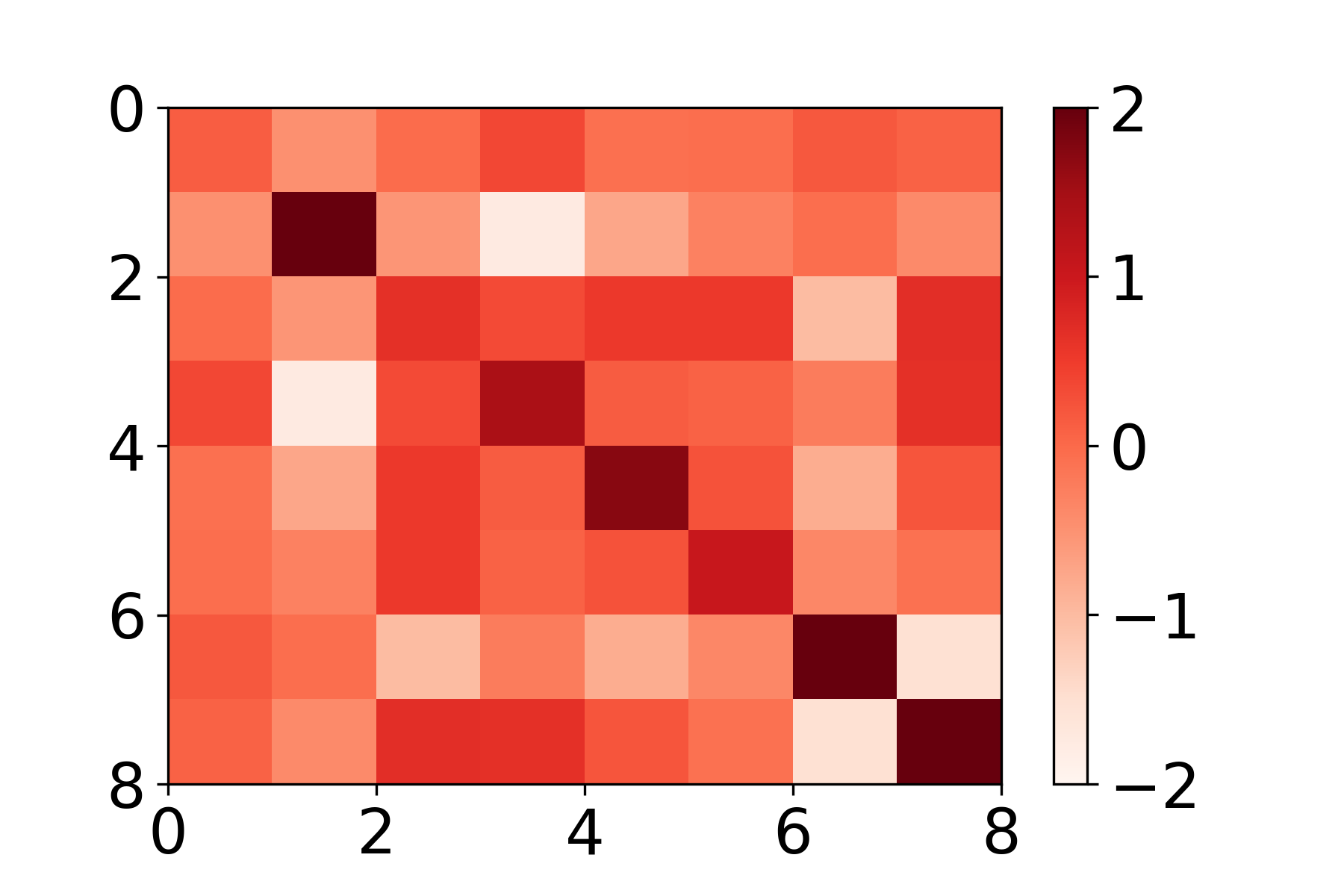}
			\caption{$\Sigma$ in $q_{c}$}
			\label{fig:graph_coupla}
		\end{subfigure}
		\begin{subfigure}{.42\columnwidth}
			\includegraphics[width=\textwidth]{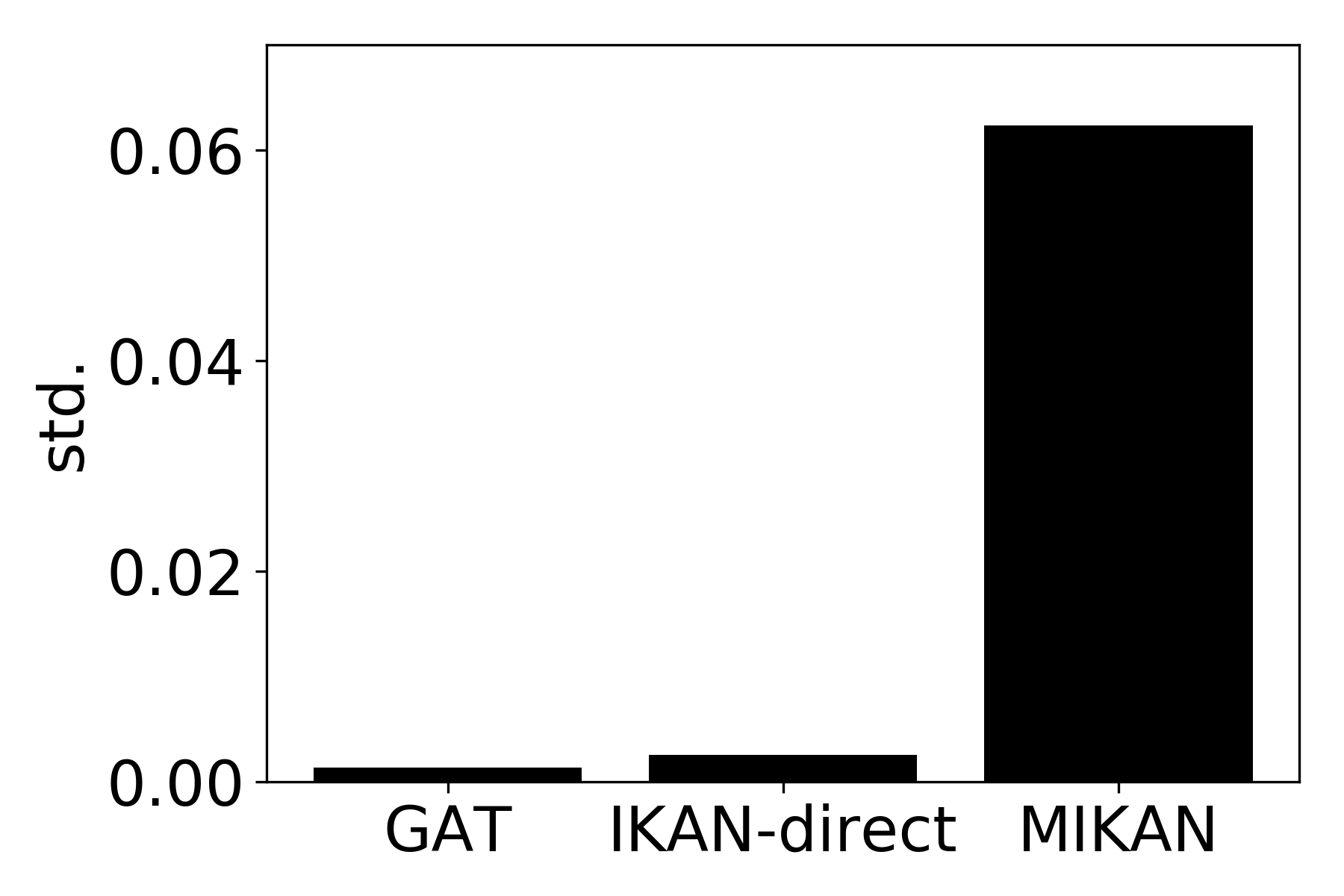}
			\caption{Std. of $\alpha_{ij}^{(m)}$}
			\label{fig:graphstdalpha}
		\end{subfigure}
		\label{fig:graph}
		\caption{(a), (b): Sampled spectral points for GAT and MIKAN. (c): Learned full covariance $\Sigma$ of $q_{c}$ in MIKAN. (d): Standard deviation of attention weights across the heads.}
	\end{figure}
	
	Figure \ref{fig:graph_rbfw} and \ref{fig:graph_mikanw1} show the spectral density in GAT and MIKAN by using t-SNE \cite{maaten2008visualizing}, respectively. The scaled dot-product attention depends on the RBF kernel, and its spectral density is fixed as a Gaussian distribution. However, MIKAN estimates the appropriate spectral density and its corresponding kernel depending on the dataset.
	Figure \ref{fig:graph_coupla} shows the learned full covariance, $\Sigma$, in $q_{c}$ of MIKAN, and Figure \ref{fig:graphstdalpha} represents the standard deviation of attention weights across the heads. A large standard deviation denotes that each head has different attention weights. The results support that MIKAN has relatively diverse attention weights across the heads by imposing the dependency between heads with the copula.
	\section{Conclusion}
	This work provides the new interpretation of the attention as a product of similarity with the RBF kernel and the magnitude with the exponential of $L^{2}$ norm. We analyze the property of the kernel and the norm in the attention, theoretically and empirically. From the derivation and analysis, we generalize the attention with an implicit kernel function and a $L^{p}$ norm. Furthermore, we propose the copula-augmented spectral density estimation for the dependence modeling in MHA to capture the diverse context. Our generalized attention can substitute the original attention in Transformer and GAT with the same algorithmic time complexity, and our experiments show better performance from our generalized attention.
	
	\section{Ethical Impact}
	We generalize the attention in Transformer and GAT, and we believe that our model is useful for capturing the underlying pattern or context of the given dataset. We expect that our proposed models contribute to both machine learning and social science.
	For the machine learning society, attention is widely used in many domains, such as natural language processing and vision. We propose a new direction to improve and interpret the attention in Transformer and GAT. 
	For the social science aspect, we can apply our model to the recommendation, psychotherapy, and political ideal point estimation of legislators by capturing the context or patterns of user log history.
	We honor the AAAI Publications Ethics and Malpractice Statement, as well as the AAAI Code of Professional Conduct. 
	\section{Acknowledgments}
	This research was supported by Basic Science Research Program through the National Research Foundation of Korea(NRF) funded by the Ministry of Education(NRF-2018R1C1B600865213)
	\bibliography{ref_ika}

\begin{thebibliography}{34}
\providecommand{\natexlab}[1]{#1}
\providecommand{\url}[1]{\texttt{#1}}
\providecommand{\urlprefix}{URL }
\expandafter\ifx\csname urlstyle\endcsname\relax
  \providecommand{\doi}[1]{doi:\discretionary{}{}{}#1}\else
  \providecommand{\doi}{doi:\discretionary{}{}{}\begingroup
  \urlstyle{rm}\Url}\fi

\bibitem[{Bahdanau et~al.(2017)Bahdanau, Brakel, Xu, Goyal, Lowe, Pineau,
  Courville, and Bengio}]{BahdanauBXGLPCB17}
Bahdanau, D.; Brakel, P.; Xu, K.; Goyal, A.; Lowe, R.; Pineau, J.; Courville,
  A.~C.; and Bengio, Y. 2017.
\newblock An Actor-Critic Algorithm for Sequence Prediction.
\newblock In \emph{5th International Conference on Learning Representations,
  {ICLR} 2017, Toulon, France, April 24-26, 2017, Conference Track
  Proceedings}. OpenReview.net.

\bibitem[{Bahdanau, Cho, and Bengio(2014)}]{bahdanau2014neural}
Bahdanau, D.; Cho, K.; and Bengio, Y. 2014.
\newblock Neural machine translation by jointly learning to align and
  translate.
\newblock \emph{arXiv preprint arXiv:1409.0473} .

\bibitem[{Chen et~al.(2018)Chen, Klushyn, Kurle, Jiang, Bayer, and
  Smagt}]{chen2018metrics}
Chen, N.; Klushyn, A.; Kurle, R.; Jiang, X.; Bayer, J.; and Smagt, P. 2018.
\newblock Metrics for deep generative models.
\newblock In \emph{International Conference on Artificial Intelligence and
  Statistics}, 1540--1550.

\bibitem[{Choromanski et~al.(2020{\natexlab{a}})Choromanski, Likhosherstov,
  Dohan, Song, Davis, Sarlos, Belanger, Colwell, and
  Weller}]{choromanski2020masked}
Choromanski, K.; Likhosherstov, V.; Dohan, D.; Song, X.; Davis, J.; Sarlos, T.;
  Belanger, D.; Colwell, L.; and Weller, A. 2020{\natexlab{a}}.
\newblock Masked Language Modeling for Proteins via Linearly Scalable
  Long-Context Transformers.
\newblock \emph{arXiv preprint arXiv:2006.03555} .

\bibitem[{Choromanski et~al.(2020{\natexlab{b}})Choromanski, Likhosherstov,
  Dohan, Song, Gane, Sarlos, Hawkins, Davis, Mohiuddin, Kaiser
  et~al.}]{choromanski2020rethinking}
Choromanski, K.; Likhosherstov, V.; Dohan, D.; Song, X.; Gane, A.; Sarlos, T.;
  Hawkins, P.; Davis, J.; Mohiuddin, A.; Kaiser, L.; et~al. 2020{\natexlab{b}}.
\newblock Rethinking attention with performers.
\newblock \emph{arXiv preprint arXiv:2009.14794} .

\bibitem[{Defferrard, Bresson, and
  Vandergheynst(2016)}]{defferrard2016convolutional}
Defferrard, M.; Bresson, X.; and Vandergheynst, P. 2016.
\newblock Convolutional neural networks on graphs with fast localized spectral
  filtering.
\newblock In \emph{Advances in neural information processing systems},
  3844--3852.

\bibitem[{Deng et~al.(2018)Deng, Kim, Chiu, Guo, and Rush}]{deng2018latent}
Deng, Y.; Kim, Y.; Chiu, J.; Guo, D.; and Rush, A. 2018.
\newblock Latent alignment and variational attention.
\newblock In \emph{Advances in Neural Information Processing Systems},
  9712--9724.

\bibitem[{Edunov et~al.(2018)Edunov, Ott, Auli, Grangier, and
  Ranzato}]{edunov2018classical}
Edunov, S.; Ott, M.; Auli, M.; Grangier, D.; and Ranzato, M. 2018.
\newblock Classical Structured Prediction Losses for Sequence to Sequence
  Learning.
\newblock In \emph{Proceedings of NAACL-HLT}, 355--364.

\bibitem[{Geetor and Lu(2003)}]{lu2003link}
Geetor, L.; and Lu, Q. 2003.
\newblock Link-based classification.
\newblock In \emph{Twelth International Conference on Machine Learning (ICML
  2003), Washington DC}.

\bibitem[{Goodfellow et~al.(2014)Goodfellow, Pouget-Abadie, Mirza, Xu,
  Warde-Farley, Ozair, Courville, and Bengio}]{goodfellow2014generative}
Goodfellow, I.; Pouget-Abadie, J.; Mirza, M.; Xu, B.; Warde-Farley, D.; Ozair,
  S.; Courville, A.; and Bengio, Y. 2014.
\newblock Generative adversarial nets.
\newblock In \emph{Advances in neural information processing systems},
  2672--2680.

\bibitem[{Jang, Gu, and Poole(2017)}]{jang2017categorical}
Jang, E.; Gu, S.; and Poole, B. 2017.
\newblock Categorical Reparametrization with Gumble-Softmax.
\newblock In \emph{International Conference on Learning Representations (ICLR
  2017)}. OpenReview. net.

\bibitem[{Jung, Song, and Park(2020)}]{jung2020approximate}
Jung, Y.; Song, K.; and Park, J. 2020.
\newblock Approximate Inference for Spectral Mixture Kernel.
\newblock \emph{arXiv preprint arXiv:2006.07036} .

\bibitem[{Katharopoulos et~al.(2020)Katharopoulos, Vyas, Pappas, and
  Fleuret}]{katharopoulos2020transformers}
Katharopoulos, A.; Vyas, A.; Pappas, N.; and Fleuret, F. 2020.
\newblock Transformers are rnns: Fast autoregressive transformers with linear
  attention.
\newblock In \emph{International Conference on Machine Learning}, 5156--5165.
  PMLR.

\bibitem[{Kim(2014)}]{kim2014convolutional}
Kim, Y. 2014.
\newblock Convolutional Neural Networks for Sentence Classification.
\newblock In \emph{Proceedings of the 2014 Conference on Empirical Methods in
  Natural Language Processing (EMNLP)}, 1746--1751.

\bibitem[{Kingma and Welling(2014)}]{KingmaW13}
Kingma, D.~P.; and Welling, M. 2014.
\newblock Auto-Encoding Variational Bayes.
\newblock In Bengio, Y.; and LeCun, Y., eds., \emph{2nd International
  Conference on Learning Representations, {ICLR} 2014, Banff, AB, Canada, April
  14-16, 2014, Conference Track Proceedings}.

\bibitem[{Kipf and Welling(2017)}]{gcnKipfW17}
Kipf, T.~N.; and Welling, M. 2017.
\newblock Semi-Supervised Classification with Graph Convolutional Networks.
\newblock In \emph{5th International Conference on Learning Representations,
  {ICLR} 2017, Toulon, France, April 24-26, 2017, Conference Track
  Proceedings}. OpenReview.net.

\bibitem[{Li et~al.(2019)Li, Chang, Mroueh, Yang, and Poczos}]{ikl2019}
Li, C.-L.; Chang, W.-C.; Mroueh, Y.; Yang, Y.; and Poczos, B. 2019.
\newblock Implicit Kernel Learning.
\newblock In \emph{The 22nd International Conference on Artificial Intelligence
  and Statistics}, 2007--2016.

\bibitem[{Li and Dunson(2019)}]{li2019geodesic}
Li, D.; and Dunson, D.~B. 2019.
\newblock Geodesic distance estimation with spherelets.
\newblock \emph{arXiv preprint arXiv:1907.00296} .

\bibitem[{Maas, Hannun, and Ng(2013)}]{maas2013rectifier}
Maas, A.~L.; Hannun, A.~Y.; and Ng, A.~Y. 2013.
\newblock Rectifier nonlinearities improve neural network acoustic models.
\newblock In \emph{in ICML Workshop on Deep Learning for Audio, Speech and
  Language Processing}. Citeseer.

\bibitem[{Maaten and Hinton(2008)}]{maaten2008visualizing}
Maaten, L. v.~d.; and Hinton, G. 2008.
\newblock Visualizing data using t-SNE.
\newblock \emph{Journal of machine learning research} 9(Nov): 2579--2605.

\bibitem[{Nelsen(2007)}]{copula2007nelsen}
Nelsen, R.~B. 2007.
\newblock \emph{An introduction to copulas}.
\newblock Springer Science \& Business Media.

\bibitem[{Perozzi, Al-Rfou, and Skiena(2014)}]{perozzi2014deepwalk}
Perozzi, B.; Al-Rfou, R.; and Skiena, S. 2014.
\newblock Deepwalk: Online learning of social representations.
\newblock In \emph{Proceedings of the 20th ACM SIGKDD international conference
  on Knowledge discovery and data mining}, 701--710.

\bibitem[{Rahimi and Recht(2008)}]{rahimi2008random}
Rahimi, A.; and Recht, B. 2008.
\newblock Random features for large-scale kernel machines.
\newblock In \emph{Advances in neural information processing systems},
  1177--1184.

\bibitem[{Reed and Simon(1975)}]{bochner}
Reed, M.; and Simon, B. 1975.
\newblock \emph{II: Fourier Analysis, Self-Adjointness}, volume~2.
\newblock Elsevier.

\bibitem[{Sklar(1959)}]{sklar1959functions}
Sklar, M. 1959.
\newblock Distribution functions in dimensions and their margins.
\newblock \emph{Publ. inst. statist. univ. Paris} 8: 229--231.

\bibitem[{Tompkins et~al.(2019)Tompkins, Senanayake, Morere, and
  Ramos}]{tompkins2019black}
Tompkins, A.; Senanayake, R.; Morere, P.; and Ramos, F. 2019.
\newblock Black Box Quantiles for Kernel Learning.
\newblock In \emph{The 22nd International Conference on Artificial Intelligence
  and Statistics}, 1427--1437.

\bibitem[{Ton et~al.(2018)Ton, Flaxman, Sejdinovic, and Bhatt}]{ton2018spatial}
Ton, J.-F.; Flaxman, S.; Sejdinovic, D.; and Bhatt, S. 2018.
\newblock Spatial mapping with Gaussian processes and nonstationary Fourier
  features.
\newblock \emph{Spatial statistics} 28: 59--78.

\bibitem[{Tran, Blei, and Airoldi(2015)}]{tran2015copula}
Tran, D.; Blei, D.; and Airoldi, E.~M. 2015.
\newblock Copula variational inference.
\newblock In \emph{Advances in Neural Information Processing Systems},
  3564--3572.

\bibitem[{Tsai et~al.(2019)Tsai, Bai, Yamada, Morency, and
  Salakhutdinov}]{tsai2019transformer}
Tsai, Y.-H.~H.; Bai, S.; Yamada, M.; Morency, L.-P.; and Salakhutdinov, R.
  2019.
\newblock Transformer Dissection: An Unified Understanding for Transformer’s
  Attention via the Lens of Kernel.
\newblock In \emph{Proceedings of the 2019 Conference on Empirical Methods in
  Natural Language Processing and the 9th International Joint Conference on
  Natural Language Processing (EMNLP-IJCNLP)}, 4335--4344.

\bibitem[{Vaswani et~al.(2017)Vaswani, Shazeer, Parmar, Uszkoreit, Jones,
  Gomez, Kaiser, and Polosukhin}]{vaswani2017attention}
Vaswani, A.; Shazeer, N.; Parmar, N.; Uszkoreit, J.; Jones, L.; Gomez, A.~N.;
  Kaiser, {\L}.; and Polosukhin, I. 2017.
\newblock Attention is all you need.
\newblock In \emph{Advances in neural information processing systems},
  5998--6008.

\bibitem[{Veli{\v{c}}kovi{\'c} et~al.(2018)Veli{\v{c}}kovi{\'c}, Cucurull,
  Casanova, Romero, Li{\`o}, and Bengio}]{Petar2018gat}
Veli{\v{c}}kovi{\'c}, P.; Cucurull, G.; Casanova, A.; Romero, A.; Li{\`o}, P.;
  and Bengio, Y. 2018.
\newblock Graph Attention Networks.
\newblock In \emph{International Conference on Learning Representations}.

\bibitem[{Wang et~al.(2020)Wang, Zhao, Lioma, Li, Zhang, and
  Simonsen}]{complexword20}
Wang, B.; Zhao, D.; Lioma, C.; Li, Q.; Zhang, P.; and Simonsen, J.~G. 2020.
\newblock Encoding word order in complex embeddings.
\newblock In \emph{8th International Conference on Learning Representations,
  {ICLR} 2020, Addis Ababa, Ethiopia, April 26-30, 2020}. OpenReview.net.

\bibitem[{Wiseman and Rush(2016)}]{wiseman2016sequence}
Wiseman, S.; and Rush, A.~M. 2016.
\newblock Sequence-to-Sequence Learning as Beam-Search Optimization.
\newblock In \emph{Proceedings of the 2016 Conference on Empirical Methods in
  Natural Language Processing}, 1296--1306.

\bibitem[{Yaglom(1987)}]{yaglom}
Yaglom, A.~M. 1987.
\newblock Correlation Theory of Stationary and Related Random Functions.
\newblock \emph{Volume I: Basic Results.} 526.

\end{thebibliography}
	\bibliographystyle{aaai}
\end{document}